\documentclass[runningheads]{llncs}
\usepackage{subcaption} 
\usepackage[T1]{fontenc}
\usepackage{graphicx}
\usepackage{booktabs}

\usepackage[misc]{ifsym}
\newcommand{\corr}{(\Letter)}

\usepackage{mwe}

\usepackage{amsmath}
\usepackage{amssymb}
\usepackage{mathtools}
\usepackage{comment}

\newtheorem{assumption}{Assumption}

\def\R{{\mathbb{R}}}

\renewcommand{\Pr}[1]{{\mathbf{Pr}\left[#1\right]}}

\def\vecpi{\boldsymbol{\pi}}

\renewcommand{\vec}[1]{\mathbf{#1}}

\begin{document}

\title{Reinforcement Learning in Switching Non-Stationary Markov Decision Processes: Algorithms and Convergence Analysis\thanks{A shortened version of this paper has been accepted for publication in the Proceedings of the 2026 IEEE Conference on Decision and Control (CDC). The present manuscript is the extended version and contains complete proofs, supporting background material, and full experimental details.}}

\titlerunning{RL in SNS-MDP: Algorithms and Convergence Analysis}

\author{Mohsen Amiri \inst{1} \orcidID{0000-0003-4704-8848} \corr 
\and Sindri Magn\'usson \inst{1} \orcidID{0000-0002-6617-8683}}


\authorrunning{M. Amiri and S. Magn\'usson}

\institute{Department of Computer and System Science, Stockholm University, SE-164 25 Stockholm, Sweden \email{\{mohsen.amiri,  sindri.magnusson\}@dsv.su.se}}

\maketitle              

\begin{abstract}
We introduce the Switching Non-Stationary Markov Decision Process (SNS-MDP) framework, in which the environment transitions among a finite set of MDPs governed by a latent Markov chain while the agent observes only the external state. We show that the long-term effect of this switching is equivalent to stationary dynamics parameterized by the stationary distribution of the hidden Markov chain. For fixed policies, we derive a closed-form expression for the SNS value function and prove that standard temporal-difference (TD) learning converges to it almost surely despite persistent non-stationarity. We further establish that policy iteration converges to the optimal policy of the equivalent averaged environment, and prove that tabular Q-learning converges almost surely to the optimal Q-function. The framework is validated on a wireless communication network with Markovian channel noise, demonstrating its practical efficacy for decision-making in rapidly time-varying systems.

\end{abstract}

\section{Introduction}
Reinforcement Learning (RL) is a powerful framework for training agents to make sequential decisions by learning from interactions with their environment. In standard RL settings, the environment is often assumed to be stationary, meaning that the transition dynamics and reward functions remain unchanged over time. This assumption simplifies analysis and algorithm design, allowing for the application of well-established techniques for policy optimization and convergence guarantees. However, many real-world problems are  non-stationary, where the environment's dynamics and reward structures evolve over time, potentially changing at every iteration. 

Non-stationarity in reinforcement learning introduces significant challenges, as the evolving nature of the environment can make it difficult for agents to maintain effective policies over time. If the environment changes in an unconstrained or arbitrary manner, it becomes impossible for the agent to learn an optimal policy, or even estimate the value of a policy, since past experiences may no longer be relevant for future decision-making. To enable learning in non-stationary settings, either the changes in the environment must occur gradually, allowing the agent enough time to adapt to the new dynamics, or there must be some underlying structure in the non-stationarity that can be exploited. 


In this paper, we propose a novel structured form of non-stationarity that both realistically models real-world challenges and can be exploited for analysis and algorithm development. Specifically, we introduce the framework of SNS-MDPs. In this setting, the environment can change at each iteration, switching among a finite set of distinct environments, each characterized by its own transition probabilities and reward functions. The differences between these environments can be arbitrarily large, capturing a wide range of scenarios. A key feature is that the agent does not know, and cannot measure or observe which environment it is currently in. Instead, the switches between environments follow a Markov chain, providing a systematic way to model the transitions and allowing for more tractable analysis and algorithm design. 

This structure captures many real-world scenarios where the underlying conditions change based on recent history, even though the agent cannot directly observe the current environment. For instance, in communication networks, the quality of the network can shift between different modes, such as high congestion during peak hours and smoother operation during off-peak hours, depending on factors like time of day and recent traffic patterns \cite{bolch2006queueing,pardoux2008markov}. Although the agent  does not know the exact congestion state, the likelihood of changes in network conditions follows a predictable pattern based on prior states, making the transitions Markovian. Similarly, in financial markets, shifts between regimes of low and high volatility or bull and bear markets occur in response to economic indicators, recent trends, or market events \cite{pardoux2008markov,choji2013markov}. While an investor cannot observe the true state of the market regime directly, the changes exhibit a form of structure that depends on recent conditions, following a Markov process.
The main contributions of this paper are as follows:
\begin{enumerate}
    \item  We introduce the novel framework of SNS-MDP, which models non-stationary environments by allowing the underlying dynamics and rewards to change according to a Markov chain.

    \item  For the case of fixed policies or Markov Reward Processes (MRPs), we define an SNS value function that remains invariant to the environmental state and show that it has a closed-form expression determined by the statistical properties of the Markov chain.

   \item  We prove that, despite the non-stationarity, TD-learning algorithms converge with probability one to the SNS value function defined in bullet 2 under a fixed policy.

    \item We demonstrate that policy improvement can be implemented within this framework, and prove that the policy iteration algorithm converges to the optimal policy for SNS-MDPs.

   \item  We prove that, even in the presence of non-stationarity,  Q-learning  converge probability one  to an on optimal SNS-MDP Q-table, provided a properly fixed behavioral policy is used.

    \item  Finally, we illustrate the practicality of SNS-MDPs through an example in communication networks, where channel noise follows a Markov chain, demonstrating the framework’s effectiveness in optimizing decision-making in non-stationary settings.
\end{enumerate}

A shortened version of this paper has been accepted for publication at the 2026 IEEE Conference on Decision and Control (CDC)~\cite{amiri2026snsmdpcdc}. This manuscript is the extended version, containing the complete proofs of all results, the supporting background on Markov chains and MRPs, and the full experimental details in the supplementary sections.

\section{Related Work}
Reinforcement learning in non-stationary Markov Decision Processes (MDPs) has been explored in previous research. Here, we review the most relevant studies and approaches that relate to our work.

Partially Observable Markov Decision Processes (POMDPs) involve scenarios where the agent cannot fully observe the underlying state~\cite{aastrom1965optimal,krishnamurthy2016partially}. 
Although the core dynamics may be stationary, non-stationarity can arise through variations in the observable components, which provide only partial information about the true state of the environment.
 While POMDPs share some similarities with our SNS-MDPs, they typically rely on the agent's ability to use observable information to infer the hidden states and adapt accordingly. In contrast, our work diverges from this paradigm by considering scenarios where the agent cannot infer the latent modes of the environment, making it necessary to develop alternative strategies for dealing with evolving dynamics without assuming access to a structured latent representation.

Another line of research in reinforcement learning focuses on non-stationary environments where the dynamics and rewards can change freely over time, with the impact of these changes reflected in the regret~\cite{auer2008near,fei2020dynamic,zhong2021optimistic,model_free_non_stationary_rl,non_stationary_linear_mdps,zhou2022nonstationary,adversarial_linear_mixture_mdps,kernel_based_nonstationary_rl}. Specifically, the regret is often bounded by the total variation in the transition probabilities and rewards across different MDPs. While these approaches are valuable, they differ significantly from our model due to the lack of structure in the changes; the MDPs can evolve arbitrarily, leading to increased regret. In contrast, our work on SNS-MDPs assumes that changes in the environment follow a Markov chain, which allows us to study the convergence of value and Q-functions under a fixed policy and to characterize these functions based on the statistical properties of the environmental Markov chain. Such analysis is not feasible with more unstructured changes, where value functions may not converge, although regret can still be bounded by the total variation. Additionally, these prior works typically address episodic tasks, whereas our focus is on continuing (infinite horizon) tasks.


Meta-RL and multi-task RL address non-stationarity by learning strategies for rapid adaptation across a distribution of tasks, typically assuming episodic settings~\cite{finn2017model,yu2020meta,xie2021deep,yu2020gradient,sodhani2021multi}. Their goal is to optimize for quick adaptation based on prior task experience, focusing on task-specific adaptation rather than long-term dynamics. In contrast, our setup is fundamentally different, as we focus on continual tasks where the transitions can change at each time step, not just between tasks, requiring the agent to adapt continuously to evolving dynamics.

The most relevant papers to our work are probably~\cite{tennenholtz2023reinforcement,guo2024average,chen2022adaptive,amiri2024convergence}. Specifically, the study in~\cite{tennenholtz2023reinforcement}, where their term "context" aligns with what we refer to as the "environment," differs primarily in terms of the observability of this context. Indeed, they assume that the context is known to the agent, is influenced by the algorithm's history, and can be directly incorporated into decision-making. This assumption represents a fundamental distinction from our work. On the other hand, the study in~\cite{guo2024average} assumes full knowledge of the MDP dynamics and rewards, focusing on average reward MDPs. Consequently, this differs from our setting, where such information is unavailable, and the agent must learn and make decisions under uncertainty. The key difference between our work and~\cite{chen2022adaptive} lies in their assumption that certain information about the context (environment) is available, such as partial knowledge of the environment and the segment length (the duration for which the context remains constant), which they use to infer the latent state of the context. However, this information may not always be accessible, especially when it is only available during training. In contrast, our framework assumes that the context is entirely unobservable, with no direct or indirect access to it. Furthermore, while~\cite{chen2022adaptive} considers the context to either change abruptly or remain constant for a known number of time steps, occurring probabilistically, we model context transitions using a Markov chain, where each state can persist or transition based on predefined probabilities. Thus, their framework can be viewed as a special case of the more general Markov chain model used in our work. In~\cite{amiri2024convergence}, the setup is similar to ours, particularly in considering a scenario where the reward function changes at each iteration, though the transition probabilities are stationary. Therefore, this scenario is a special case of our work. Generally, in MDPs, dealing with changing reward functions is more straightforward than handling varying transition probabilities, as the latter directly alters the stochastic process. Moreover, ~\cite{amiri2024convergence} is limited to policy evaluation and does not consider other RL tasks.

\section{Notation}

We represent non-random vectors using lowercase bold letters and non-random matrices using uppercase bold letters. For example, a vector $\mathbf{v} \in \mathbb{R}^n$ and a matrix $\mathbf{A} \in \mathbb{R}^{n \times m}$ are typical representations. The expression \(\mathbf{v}(i)\) indicates the \(i\)-th component of the vector \(\mathbf{v}\), and \(\mathbf{A}(i, j)\) refers to the element located at the \(i\)-th row and \(j\)-th column of the matrix \(\mathbf{A}\). For vectors, \(\mathbf{x} \in \mathbb{R}^n\), the function \(\texttt{Diag}(\mathbf{x})\) denotes the \(n \times n\) diagonal matrix with the elements of \(\mathbf{x}\) along its diagonal.
Sets are denoted using a calligraphic typeface. We use the notation \(X \sim p(\cdot)\) to denote that \(X\) is a random variable sampled from the probability distribution \(p(\cdot)\). The probability of \(X\) being in  \(\mathcal{X}\) is expressed as \(\Pr{X \in \mathcal{X}}\).


\section{Markov Decision Processes (MDPs)} \label{Section:Preliminaries}

This paper considers reinforcement learning algorithms in MDPs. A stationary MDP is defined by the tuple $(\mathcal{S}, \mathcal{A}, p(\cdot), \vec{r}(\cdot), \gamma)$, where $\mathcal{S}$ denotes the set of states, $\mathcal{A}$ is the set of actions, $p(s' \mid s, a)$ represents the transition probability of moving to state $s'$ from state $s$ after taking action $a$, $\vec{r}(s, a)$ is the reward received in state $s$ and action $a$, and $\gamma \in [0, 1)$ is the discount factor that determines the importance of future rewards. 

A policy $\mu: \mathcal{S} \rightarrow \Delta(\mathcal{A})$ defines a distribution over the action set $\mathcal{A}$ for a given state $s$, where $\mu(a \mid s)$ specifies the probability of taking action $a$ in state $s$. 
The agent’s interaction with the environment produces a sequence of states \( S_k \), actions \( A_k \), and reward \( R_k = \vec{r}(S_k, A_k) \). The value function of a policy \(\mu\), denoted as \(\vec{v}^\mu(s)\), describes the expected cumulative discounted reward when starting from a state \(s\) and following the policy \(\mu\):

\[
\vec{v}^\mu(s) = \mathbb{E}\left[ \sum_{k=0}^{\infty} \gamma^k R_k \,\middle|\, S_0 = s \right].
\]

Similarly, the Q-function of a policy \(\mu\) (the behavior policy), denoted as \(\vec{Q}^\mu(s, a)\):

\[
\vec{Q}^\mu(s, a) = \mathbb{E}\left[ \sum_{k=0}^{\infty} \gamma^k R_k \,\middle|\, S_0 = s, A_0 = a \right].
\]

The value function has a closed-form solution. Specifically, defining the transition matrix \(\vec{P}^{\mu}\in \mathbb{R}^{|\mathcal{S}|\times |\mathcal{S}|}\) of the Markov chain induced by policy \(\mu\) as \(\vec{P}^{\mu}(s,s')=\sum_{a\in\mathcal{A}}p(s'|s,a)\mu(a|s)\).
Then the value of the policy $\mu$ can be expressed in closed form as:
\begin{equation} \label{eq:Value_Closed_Form}
    \vec{v}^{\mu}=(\vec{I}-\gamma \vec{P}^{\mu})^{-1} \vec{r}^{\mu},
\end{equation}
where \(\vec{I}\) is the identity matrix and \(\vec{r}^{\mu}(\cdot)\) is defined as \(\vec{r}^{\mu}(s) = \sum_{a \in \mathcal{A}} \vec{r}(s,a)\mu(a|s)\). The optimal policy \(\mu^{\star}\) maximizes the value function for all states, resulting in the optimal value function \(\vec{v}^{\star}(s) = \max_{\mu}\vec{v}^{\mu}(s)\).

Among the key tasks in reinforcement learning are policy evaluation, which involves estimating the value function \(\vec{v}^{\mu}\) for a given policy \(\mu\), and policy iteration, which aims to find an optimal policy by iteratively performing policy evaluation followed by policy improvement to reach the optimal policy \(\mu^{\star}\). Additionally, off-policy learning methods, such as Q-learning, play an important role, as they enable learning about one policy (the target policy) while following a different policy (the behavior policy) to collect data. These tasks are well-established in the context of stationary MDPs. However, in non-stationary MDPs, where the transition probabilities $p_k(s'|s,a)$ and rewards $\vec{r}_k(s,a)$ change at each time step $k$, the algorithms may not converge, especially if the dynamics change too rapidly or in an unconstrained manner. Even when they do converge, it is often unclear to what solution they converge. Without additional structure on the non-stationary MDP, reliable convergence is not guaranteed.

We provide one such structure on the non-stationarity that is useful for modeling practical problems and introduces regularities that can be leveraged to analyze and understand the convergence behavior of RL algorithms.

\section{Switching Non-Stationary Markov Decision Process}

In many real-world decision-making problems, the environment evolves over time, making stationary MDPs inadequate for capturing the complexity of these systems. For example, in autonomous driving, traffic conditions, such as congestion, weather, and road closures, may change in ways that affect optimal decision-making. Similarly, in communication networks, transmission quality can shift due to interference, signal degradation, or network congestion, all of which affect how agents should adapt their strategies. In these settings, the agent must make decisions without directly knowing the underlying state of the environment, which switches dynamically between different regimes.

 
We introduce SNS-MDP, a non-stationary MDP where the environment alternates between multiple latent states (or "modes"). Crucially, the agent cannot observe or measure the current latent mode and must make decisions solely based on its direct interactions with the environment. The switching between environments is governed by a Markov chain, meaning that the environment transitions probabilistically between modes depending on the current mode, though the agent remains unaware of these transitions.

Formally, SNS-MDPs are defined over a state space $\mathcal{S} = \{1, \dots, |\mathcal{S}|\}$ and an action space $\mathcal{A} = \{1, \dots, |\mathcal{A}|\}$, similar to traditional MDPs. However, unlike stationary MDPs, the environment, specifically the transition probabilities and rewards, changes at each time step. The environment can be in one of a finite number of environmental states, represented by the set $\mathcal{E}=\{1,\dots, |\mathcal{E}|\}$, where each state corresponds to a distinct configuration of the system. Each environmental state $e \in \mathcal{E}$ is associated with a unique transition probability function $p_e: \mathcal{S} \times \mathcal{A} \times \mathcal{S} \rightarrow [0,1]$ and a reward function $\vec{r}_e: \mathcal{S} \times \mathcal{A} \rightarrow \mathbb{R}$.


The dynamics of the environmental states is captured by a Markov chain $(\mathcal{E}, q(\cdot))$, where $\mathcal{E}$ denotes the set of environment states, and $q(\cdot)$ defines the transition probabilities between them, formalized in this definition. 

\begin{definition}
\label{definition:SNS-MDP}
  An \textbf{SNS-MDP} is a tuple $(\mathcal{S}, \mathcal{A}, (p_e(\cdot))_{e \in \mathcal{E}}, (\vec{r}_e(\cdot))_{e \in \mathcal{E}}, \gamma; \mathcal{E}, q(\cdot))$, where $(\mathcal{E}, q(\cdot))$ is a Markov chain over environmental states $\mathcal{E}$ and each configuration $(\mathcal{S},\mathcal{A}, p_e(\cdot), \vec{r}_e(\cdot), \gamma)$, for all $e\in \mathcal{E}$, represents a Markov Decision Process.


 %
\end{definition}


Given a realization of an SNS-MDP, we get a trajectory of the measurable states, actions, and rewards:
\begin{align} \label{eq:ORtrajectory_MDP} 
S_0, A_0, R_0, S_1, A_1, R_1, \ldots, S_k, A_k, R_k, \ldots,
\end{align}
where the reward is \( R_k = \vec{r}_{E_k}(S_k, A_k) \). At the same time, the unmeasurable environmental states evolve according to the following trajectory:
\begin{align} \label{eq:Etrajectory_MDP} 
E_0, E_1, \ldots, E_k, \ldots.
\end{align}
The key point is that the environmental states determine which transition function and reward structure are applied at each time step. Specifically, if at time $k$ the system is in environmental state $E_k = e$, then the next state follows the distribution $S_{k+1} \sim p_e(\cdot ~| S_k = s, A_k = a)$, and the reward is given by $\vec{r}_{E_k}(S_k, A_k)$. However, since the environmental state is unmeasurable, the agent must act without direct knowledge of $E_k$, relying only on the observable state $S_k$ and the history of its interactions.

This type of non-stationarity appears in many real-world applications. Consider, for example, wireless communication. At each time step, a transmitting node must decide on a communication protocol (the action) to maximize data throughput or minimize latency. The choice of protocol can include options like modulation schemes, power levels, or channel access methods. However, the wireless environment is non-stationary due to factors such as interference, network congestion, or signal fading. These factors represent the unmeasurable environmental states that influence both the success of the transmission and the quality of the communication link. Importantly, these environmental factors often follow Markovian dynamics, which means that they evolve according to a Markov chain, as modeled by $(\mathcal{E},q(\cdot))$.  While the transmitting node cannot mesure or observe the environmental state $E_k$, it must still adapt its actions based on observable system states and past experience.

\subsection{Impossibility of Inferring the Latent Environmental State}

Although the SNS-MDP can be cast syntactically as a POMDP with hidden state $(S_k, E_k)$ and observation $S_k$, POMDP methods are inapplicable in our setting. Such methods require forming a belief over the latent state, which in turn requires knowledge of $q(e' \mid e)$ and $\{p_e(s' \mid s,a)\}_{e \in \mathcal{E}}$, neither of which is available to the agent. More fundamentally, the latent environmental state is non-identifiable: distinct latent processes and model families can generate identical distributions over observable trajectories, making recovery of $E_k$ from data ill-posed regardless of the estimation strategy. The following proposition formalizes this.

\begin{proposition}\label{prop:nonidentifiable}
Let $|\mathcal{E}| = m \ge 2$. For any estimator $\hat{E}_k = g_k(\mathcal{O}_k)$ of the latent environmental state based on the observable history $\mathcal{O}_k = (S_0, A_0, \ldots, S_k, A_k)$, there exists a model $M^{\ast}$ with unknown transitions $q(e' \mid e)$ and unknown kernels $p_e(s' \mid s,a)$ such that
\[
\Pr{\hat{E}_k = E_k} = \frac{1}{m}.
\]
Thus, no estimator performs better than random guessing in the worst case.
\end{proposition}
\begin{proof}
Fix any estimator $\hat{E}_k = g_k(\mathcal{O}_k)$ and any policy. Construct $M^{\ast}$ by letting $(E_k)_{k \ge 0}$ be i.i.d.\ uniform on $\mathcal{E}$, i.e., $q(e' \mid e) = 1/m$ and $\Pr{E_0 = e} = 1/m$, and by setting $p_e(s' \mid s,a) = p(s' \mid s,a)$ for all $e \in \mathcal{E}$, for some kernel $p$ that does not depend on $e$. Then $(S_k)_{k \ge 0}$ evolves independently of $(E_k)_{k \ge 0}$, so $\mathcal{O}_k$ is independent of $E_k$ under $M^{\ast}$. Consequently,
\begin{align*}
\Pr{\hat{E}_k = E_k}
&= \sum_{e \in \mathcal{E}} \Pr{E_k = e}\,\Pr{g_k(\mathcal{O}_k) = e} \\
&= \frac{1}{m}\sum_{e \in \mathcal{E}} \Pr{g_k(\mathcal{O}_k) = e} = \frac{1}{m},
\end{align*}
where the last equality uses that $g_k$ outputs exactly one value in $\mathcal{E}$. Hence no estimator outperforms random guessing in the worst case.
\end{proof}

Since the agent also does not know $|\mathcal{E}|$, different latent processes and mode-dependent dynamics can generate identical observed distributions. Belief-state estimation and POMDP-based RL techniques are therefore inapplicable, which motivates applying standard tabular RL algorithms directly to the observable trajectory and characterizing their limits.

\section{Policy Evaluation in SNS-MDPs} \label{sec:PE}

A central problem in reinforcement learning is policy evaluation, where the objective is to estimate the value of a given policy. Once a policy is fixed, the problem essentially reduces to determining the value in a corresponding reward process.
 Therefore, to simplify notation, we consider the reward process in this section, abstracting away actions. However, when performing policy evaluation for a specific policy, these results and algorithms directly apply to the reward process induced by that policy, we investigate this in the next section.

We consider Switching Non-Stationary Markov Reward Process (SNS-MRP) formally defined as follows.
\begin{definition}\label{definition:NS-MRP}
 An \textbf{SNS-MRP} is a tuple $(\mathcal{S}, (p_e(\cdot))_{e \in \mathcal{E}}, (\vec{r}_e(\cdot))_{e \in \mathcal{E}}, \gamma; \mathcal{E}, q(\cdot))$, where $(\mathcal{E}, q(\cdot))$ is a Markov chain over environmental states $\mathcal{E}$ and each configuration $(\mathcal{S}, p_e(\cdot), \vec{r}_e(\cdot), \gamma)$, for all $e\in \mathcal{E}$, represents a Markov Reward Process. We define the transition probability matrix $\vec{P}_e\in \R^{|\mathcal{S}|\times |\mathcal{S}|}$ for each environmental state $e\in \mathcal{E}$ as  $\vec{P}_e(s, s') = p_e(s'|s)$ and the reward matrix $\vec{R}\in \R^{|\mathcal{S}|\times |\mathcal{E}|}$ as $\vec{R}(s,e)=\vec{r}_e(s)$.
\end{definition}
We make the following assumption on the SNS-MRPs.
\begin{assumption}\label{assumption:MCs_IrredApper}
    The Markov chains \(  (\mathcal{S}, p_{\text{e}}(\cdot)) \), for all \( e \in \mathcal{E} \), and \( (\mathcal{E}, q(\cdot)) \) are irreducible and aperiodic.
\end{assumption}


Our goal is to characterize the value function of SNS-MRPs. Given a measurable trajectory $S_k, R_k$ [cf. Eq.~\eqref{eq:ORtrajectory_MDP}] and a corresponding unmeasurable trajectory of environmental states $E_k$ [cf.~\eqref{eq:Etrajectory_MDP}], a natural definition of the value function is
%
%
%
\begin{align} \label{eq:value_s_e}
\vec{v}(s, e) = \mathbb{E} \left[ \sum_{k=0}^{\infty} \gamma^k R_k \mid S_0 = s, E_0 = e \right].
\end{align}
However, since the environmental state $E_k$ is unmeasurable,  we do not know the  current value under this definition. This makes it impractical, especially, for reinforcement learning  algorithms. We need a definition of the value function that relies only on the observable state $S_k$, making it more applicable in practice.  By Assumption~\ref{assumption:MCs_IrredApper}, we know that the environment Markov chain has a stationary distribution $\vecpi_{\mathcal{E}}(\cdot)$. 
Since the stationary distribution describes the long-run behavior of $E_k$ once it has stabilized, it provides a reasonable basis for defining the value function. We thus propose the following value function for SNS-MRPs:
\begin{align} \label{eq:value_s}
\vec{v}^{\texttt{SNS}}(s) = \mathbb{E}_{E \sim \vecpi_{\mathcal{E}}(\cdot)} \left[ \vec{v}(s, E) \right],
\end{align}
where the expected value is taken over the stationary distribution $E \sim \vecpi_{\mathcal{E}}(\cdot)$.
This definition allows us to capture the expected accumulated reward based solely on the observable state $S_k$, while accounting for the environmental uncertainty through the stationary distribution.

We now demonstrate that, surprisingly, the value in Eq.~\eqref{eq:value_s} has a closed form expression that can be characterized by the statistical properties of the SNS-MRP.  This is unexpected because, although the value function in stationary MRPs has a closed-form solution, as shown in Eq.~\eqref{eq:Value_Closed_Form}, the value function in non-stationary MRPs typically does not.
\begin{theorem}\label{Thm:value_closedform}
Consider a SNS-MRP under Assumption~\ref{assumption:MCs_IrredApper}. 
 Then the value function in Equation \eqref{eq:value_s} can be expressed in closed form as follows:
\begin{equation}\label{Equ:value_closedform}
    \vec{v}^{\texttt{SNS}} = \left(\vec{I} - \gamma\left( \sum_{e\in \mathcal{E}} \vecpi_{\mathcal{E}}(e) \vec{P}_e\right)\right)^{-1} \vec{r}_{\mathcal{E}} 
\end{equation}
where $\vec{r}_{\mathcal{E}}=\vec{R} \vecpi_{\mathcal{E}}$.
\end{theorem}
\begin{proof}
    See the Supplementary Materials.
\end{proof}

The theorem establishes that the value function for the SNS-MRP has a closed-form expression. It is insightful to compare this expression with the closed-form solution for a stationary MRP. In the stationary case, given a transition matrix
$\vec{P}\in \R^{|\mathcal{S}|\times|\mathcal{S}| }$ and reward vector $\vec{r}\in \R^{|\mathcal{S}|}$ then the value function is
\begin{align} \label{eq:v_for_stationary_MRP}
     \vec{v}= (\vec{I}-\gamma \vec{P})^{-1} \vec{r}.
\end{align} 
Interestingly, the closed-form expression for the SNS-MRP has a similar structure, but with key differences in  the transition matrix and the reward vector.
\begin{enumerate}
    \item Transition Matrix: Instead of the transition matrix $\vec{P}$, the SNS-MRP involves the expression:
    \begin{align} \label{eq:}
     \sum_{e\in \mathcal{E}} \vecpi_{\mathcal{E}}(e) \vec{P}_e . 
\end{align}
This is a weighted average of the transition matrices $\vec{P}_e$ corresponding to the different environmental states $e\in \mathcal{E}$. The weights $\vecpi_{\mathcal{E}}$ are given by the stationary distribution of the underlying environmental Markov chain $(\mathcal{E},q(\cdot))$. Therefore, instead of a single transition matrix $\vec{P}$, we have a weighted combination of the transition matrices across the different environmental states.
\item Reward Vector: Similarly, the reward vector $\vec{r}$
in the stationary MRP is replaced by $\vec{r}_{\mathcal{E}}=\vec{R}\vecpi_{\mathcal{E}}$ in the SNS-MRP case. This represents the weighted mean of the rewards for the different environmental states, where the weights are again given by the stationary distribution $\vecpi_{\mathcal{E}}$.
\end{enumerate}



In reinforcement learning, we aim to estimate the reward function from data without having prior knowledge of the transition probabilities or rewards. This estimation is typically performed using observed trajectories. A common approach for this is TD-learning. However, in the case of a non-stationary environment, it is uncertain whether TD-learning will converge, or if it does, to what point it will converge, since the environment's underlying dynamics are continually changing. Nonetheless, one might implement the TD update  directly on the observed states $S_k$, adapting the learning process to the measurable components of the system. To that end, we consider the following TD-learning algorithm. At each time step $k$ we perform the TD-update
\begin{equation} \label{Equ:TDlearning-adopted}
    \vec{v}_{k+1}(s) {=} \vec{v}_k(s) {+} \alpha_k \left(R_k {+} \gamma \vec{v}_k(S_{k+1}) {-} \vec{v}_k(s)\right)
\end{equation}
if  $s = S_k$ and for all other states $s\neq S_k$, we set $\vec{v}_{k+1}(s)=\vec{v}_{k}(s)$. 
The algorithm starts with an initial value vector $\vec{v}_0 \in \mathbb{R}^{|\mathcal{S}|}$, where $\alpha_k$ is the learning rate.

Our next result demonstrates that the TD-learning algorithm in Eq.~\eqref{Equ:TDlearning-adopted} converges in probability.  Moreover, we establish that it converges specifically to the SNS-MRP value $\vec{v}^{\texttt{SNS}}$ in Eq.~\eqref{Equ:value_closedform}. 

\begin{theorem}\label{Thm:main}
Consider an SNS-MRP as defined in Definition \ref{definition:NS-MRP} and let Assumption \ref{assumption:MCs_IrredApper} hold true. Then, the TD algorithm in Equation \eqref{Equ:TDlearning-adopted}, with the step-sizes 
\begin{equation}\label{Equ:E1}
    \sum_{k=0}^{\infty} \alpha_k = \infty \quad \text{and} \quad \sum_{k=0}^{\infty} \alpha_k^2 < \infty,
\end{equation}
converges with probability one to the fixed-point \(\lim_{k \rightarrow \infty} \vec{v}_k = \vec{v}^{\texttt{SNS}}\), where \(\vec{v}^{\texttt{SNS}}\) is defined by Eq.~\eqref{Equ:value_closedform}.
\end{theorem}
\begin{proof}
 See the Supplementary Materials.
\end{proof}
The theorem ensures that, under the SNS structure, TD-learning converges to a fixed point, and this fixed point corresponds the SNS-MRP value function in Eq.~\eqref{eq:value_s}. This guarantees that, despite the non-stationarity of the environment, the algorithm reliably captures the long-term value of states as they evolve. 

\section{Policy Iteration in SNS-MDPs} \label{sec:PI}


 We now focus our attention to learning the optimal policy in SNS-MDPs. The goal is to learn the optimal policy, i.e., the one that optimizes the value function. 

 In a SNS-MDP, we must constrain ourselves to policies that are based only on the measurable states $S_k$, but not based on the environment states $E_k$, since they are not known to the agent. Therefore, we focus on policies of the form $\mu:\mathcal{S}\rightarrow \Delta(\mathcal{A})$,  where the policy $\mu$ maps each state $S_k$ to a probability distribution over actions, without relying on the unknown environmental state $E_k$.
 We denote the probability of selecting action $a$ given state $s$ under policy $\mu$ as $\mu(a|s)$.


When searching for the optimal policy, it is often helpful to consider the state-action value function, or Q-function. Given a policy $\mu$, the Q-function is 
\begin{align*} 
\vec{Q}^{\mu}(s, e, a) = \mathbb{E} \left[ \sum_{k=0}^{\infty} \gamma^k R_k \,\middle|\, S_0 = s,\, E_0 = e,\, A_0 = a \right].
\end{align*}
 However, since the environmental state $E_k$ is unmeasurable, we cannot directly condition the action-value function on it. Instead, we must rely on a Q-function that depends solely on the observable state $S_k$ and the actions taken, ignoring any direct information about the underlying environmental state. Similarly as before, we consider the expected value taken over the stationary distribution $E\sim \vecpi(\cdot)$. In particular, we consider the following state-action value:
\begin{align} 
\vec{Q}^{\texttt{SNS},\mu}(s, a) = 
 \mathbb{E}_{E \sim \pi_{\mathcal{E}}(\cdot)} \left[ \vec{Q}^{\mu}(s, E, a) \right]. \label{eq:Qvalue_s_a}
\end{align}
We can connect this Q-table to the SNS-MRP value function associated with the policy \(\mu\); specifically, for each environment state \( e \in \mathcal{E} \), we define the transition matrix \(\vec{P}_e^{\mu} \in \mathbb{R}^{|\mathcal{S}|\times |\mathcal{S}|}\) by \(\vec{P}_e^{\mu}(s,s') = \sum_{a \in \mathcal{A}} p_e(s'|s,a)\mu(a|s)\).
 Then, by Theorem~\ref{Thm:value_closedform}, the SNS-MRP value  under the fixed policy $\mu$ is
 $$\vec{v}^{\texttt{SNS},\mu} = \left(\vec{I} - \gamma\left( \sum_{e\in \mathcal{E}} \vecpi_{\mathcal{E}}(e) \vec{P}_e^{\mu}\right)\right)^{-1} \vec{r}^{\mu}_{\mathcal{E}}$$
where \(\vec{r}^{\mu}_{\mathcal{E}} = \vec{R}^{\mu}\vec{\pi}_{\mathcal{E}}\), and \(\vec{R}^{\mu}\in\mathbb{R}^{|\mathcal{S}|\times|\mathcal{E}|}\) is the reward matrix defined as \(\vec{R}^{\mu}(s,e) = \vec{r}_{e}^{\mu}(s) = \sum_{a \in \mathcal{A}}\vec{r}_e(s,a)\mu(a|s)\). We can now establish the relationship between the SNS value function $\vec{v}^{\texttt{SNS},\mu}$ and the SNS Q-function $\vec{Q}^{\texttt{SNS},\mu}$ as follows.

%
%
%
%
\begin{lemma} \label{lemma:QFunc_ValueFunc}
    For any state-action pair $(s, a)\in \mathcal{S}\times \mathcal{A}$, the SNS Q-function and value function are related by the equation:
    \begin{align}\label{eq:QFunc_ValueFunc}
    \vec{Q}^{\texttt{SNS},\mu}(s, a) = \vec{r}_{\mathcal{E}}(s,a) + \gamma \sum_{s'\in \mathcal{S}} p(s'|s,a) \vec{v}^{\texttt{SNS},\mu}(s')
    \end{align}
    where 
    $\vec{r}_{\mathcal{E}}(s,a) = \sum_{e \in \mathcal{E}} \vec{r}_e(s,a) \vecpi_{\mathcal{E}}(e)$.

\end{lemma}
\begin{proof}
 See the Supplementary Materials.
\end{proof}

This lemma establishes the relationship between the SNS state-value function, $\vec{v}^{\texttt{SNS},\mu}$, and the SNS Q-function, $\vec{Q}^{\texttt{SNS},\mu}$. In Section~\ref{sec:PE}, we have already demonstrated how to estimate $\vec{v}^{\texttt{SNS},\mu}$. To connect it with $\vec{Q}^{\texttt{SNS},\mu}$ we  need  the transition probabilities $p_e(s'|s,a)$. In deterministic environments, such as Grid-World or shortest-path problems, these transitions are explicitly known for each action, making this connection straightforward.
In more general settings, $\vec{Q}^{\texttt{SNS},\mu}$ can still be 
estimated using TD learning for policy evaluation, similarly as in Section~\ref{sec:PE}, under the fixed policy $\mu$.



If we can estimate or recover the SNS Q-table, $\vec{Q}^{\texttt{SNS},\mu}$, then it plausible to perform Policy Iteration. 
We begin with an initial policy, $\mu^0$, and then, at each iteration $n$, estimate the SNS state-action values for the current policy $\mu^n$, i.e., $\vec{Q}^{\texttt{SNS},\mu^n}$. The policy is subsequently updated according to
\begin{align}\label{eq:monotonic_improvement_PI_eq_1}
\mu^{n+1}(s) = \arg\max_{a \in \mathcal{A}} \vec{Q}^{\texttt{SNS},\mu^n}(s, a)
\end{align}
ensuring that
\begin{align}\label{eq:monotonic_improvement_PI_eq_2}
\vec{Q}^{\texttt{SNS},\mu^n}\big(s, \mu^{n+1}(s)\big) = \max_{a \in \mathcal{A}} \vec{Q}^{\texttt{SNS},\mu^n}(s, a).
\end{align}
 We now establish that the Policy Iteration algorithm works in SNS-MDPs. 
\begin{theorem} 
\label{Thm:monotonic_improvement}
Consider two policies $\mu(\cdot)$, $\mu'(\cdot)$, and define
\[
\vec{Q}^{\texttt{SNS},\mu}(s, \mu') = \mathbb{E}_{a \sim \mu'(\cdot|s)}[\vec{Q}^{\texttt{SNS},\mu}(s, a)].
\]
If $\vec{Q}^{\texttt{SNS},\mu}(s, \mu') \geq \vec{v}^{\texttt{SNS},\mu}(s)$ for all $s\in \mathcal{S}$ then it holds that $\vec{v}^{\texttt{SNS},\mu'}(s) \geq \vec{v}^{\texttt{SNS},\mu}(s)$ for all $s\in \mathcal{S}$. 
\end{theorem}
\begin{proof}
 See the Supplementary Materials.
\end{proof}
The theorem establishes that $\mu'$ is at least as good a policy as $\mu$, ensuring that the Policy Iteration algorithm improves the policy at each step. We will now demonstrate that this improvement continues until a fixed point is reached, at which point the algorithm converges to the optimal policy.
\begin{theorem}\label{Thm:monotonic_improvement_PI}
Let \(\{\mu^n\}\) be a sequence of policies generated by the Policy Improvement algorithm in Eq.~\eqref{eq:monotonic_improvement_PI_eq_1}.
If $\mu^{n+1}=\mu^n$ for some $n$ then the policy $\mu^n$ is the optimal policy in the sense that
$$\mu^n(s)=  \underset{\mu}{\rm{argmax}} ~~\vec{v}^{\texttt{SNS},\mu}(s) ~~\text{ for all } s\in \mathcal{S}.
$$
%
\end{theorem}
\begin{proof}
 See the Supplementary Materials.
\end{proof}

\section{Q-learning in SNS-MDP}

The Policy Iteration algorithm discussed in the previous section has a drawback: at each iteration $k$, we must estimate the Q-table \(\vec{Q}^{\texttt{SNS},\mu^k}\) for the corresponding policy $\mu^k$. In contrast, Q-learning often provides a more efficient alternative, as it directly estimates the optimal Q-table without requiring explicit policy evaluation at each step. However, the convergence of Q-learning is generally not guaranteed outside of stationary environments. We now show that in SNS-MDPs, under certain conditions, Q-learning does converge to a stationary Q-table.

 In Q-learning, the goal is to learn the optimal Q-table from sampled interaction. In SNS-MDPs, we observe two types of sample trajectories: the measurable trajectory
[see Eq.~\eqref{eq:ORtrajectory_MDP}]
 $S_k,A_k,R_k$
 and the unmeasurable trajectory [see Eq.~\eqref{eq:Etrajectory_MDP}] $E_k$.
Our goal is to learn an optimal Q-table, \(\vec{Q}^{\texttt{SNS}} \in \R^{|\mathcal{S}|\times|\mathcal{A}|}\), using only the observable trajectory.
In particular, if we let \(\vec{Q}^{\texttt{SNS}}_k\) be the Q-table at iteration $k$, with \(\vec{Q}^{\texttt{SNS}}_0\) initialized arbitrarily, i.e., as a zero matrix, then we may perform the following update:
\begin{align}\label{eq:QL}
\vec{Q}^{\texttt{SNS}}_{k+1}(s, a) 
&= (1 {-} \alpha_k)\,\vec{Q}^{\texttt{SNS}}_{k}(S_k, A_k) 
{+} \alpha_k \Bigl(\vec{r}_{e}(S_k,A_k) {+} \gamma \,\max_{a \in \mathcal{A}} \vec{Q}^{\texttt{SNS}}_{k}(S_{k+1}, a)\Bigr)
\end{align}
if $s{=}S_k$ and $a{=}A_k$ and $\vec{Q}^{\texttt{SNS}}_{k+1}(s, a){=} \vec{Q}^{\texttt{SNS}}_{k}(s, a)$ otherwise. Since the environment mode changes at each iteration according to $E_k$, it is unclear whether Q-learning converges and, if so, to what value. We now establish that Q-learning does converge in SNS-MDPs. To characterize the limit of this convergence, we define
%
$\vec{v}^{\texttt{SNS},\star}(s) = \max_{\mu} \, \vec{v}^{\texttt{SNS},\mu}(s).$
%
 We then show that, under certain conditions, Q-learning converges to
\begin{align}\label{eq:QFunc_ValueFunc_optimal}
\vec{Q}^{\texttt{SNS},\star}(s, a) = \vec{r}_{\mathcal{E}}(s,a) + \gamma \sum_{s' \in \mathcal{S}} p(s' \mid s,a)\,\vec{v}^{\texttt{SNS},\star}(s'),
\end{align}
Before proving this result, we first define the optimal Bellman operator for SNS-MDPs. The following lemma establishes the uniqueness of the optimal Q-table as the fixed point of this operator.
\begin{lemma}\label{Thm:QL_uniqness}
The Q-table $\vec{Q}^{\texttt{SNS},\star}$ in Eq.~\eqref{eq:QFunc_ValueFunc_optimal}  is the unique solution to the Bellman optimality equation:
\begin{align}\label{eq:QFunc_QFunc_optimal}
\vec{Q}^{\texttt{SNS},\star}(s, a) 
= \vec{r}_{\mathcal{E}}(s,a) + \gamma \sum_{s' \in \mathcal{S}} p(s' \mid s,a) \,\max_{a' \in \mathcal{A}} \vec{Q}^{\texttt{SNS},\star}(s', a').
\end{align}
\end{lemma}
\begin{proof}
See the Supplementary Materials for details.
\end{proof}
\noindent We now establish the convergence of the Q-learning algorithm in SNS-MDPs.
\begin{theorem}\label{Thm:QL_convergence}
 %
 Suppose that the steps-sizes $\alpha_k$ satisfy the condition in Eq.~\eqref{Equ:E1} and  every combination of state \(s \in \mathcal{S}\), action \(a\in \mathcal{A}\), and environmental state \(e \in \mathcal{E}\) are visited infinitely often then the sequence $\vec{Q}_k^{\texttt{SNS}}$ converges with probability one to the fixed point $\lim_{k\rightarrow \infty} \vec{Q}_k^{\texttt{SNS}} = \vec{Q}^{\texttt{SNS},\star}$.
%
\end{theorem}
\begin{proof}
See the Supplementary Materials for details.
\end{proof}
This result establishes that, under appropriate step-size conditions and sufficient exploration, Q-learning in SNS-MDPs converges almost surely to the optimal Q-table. The requirement that every state-action-environment triplet $(s,a,e)$ is visited infinitely often ensures that the learning process adequately samples the entire state space, allowing the algorithm to correctly estimate the value function despite the underlying non-stationarity. Without this condition, the algorithm may fail to learn optimal Q-values for underexplored regions, potentially leading to suboptimal policies.

\section{Experimental Results} \label{section:ExperimentalResults}

We now demonstrate our theoretical results in the context of wireless communication systems, which often experience dynamic channel conditions due to factors such as fading, interference, and user mobility.
%
To enhance performance under such fluctuating conditions, Adaptive Modulation (AM) techniques are employed, where transmission parameters are dynamically adjusted \cite{huang2020adaptive,qiu1999performance}. To show the effectiveness of the proposed framework, we model an adaptive communication system using the SNS-MDP framework, which effectively captures the stochastic nature of wireless environments. 

We consider a scenario where the transceiver, functioning as an agent, selects a frequency band for data transmission by observing the current modulation. This selection is the agent's action, i.e., the action space is 
$\mathcal{A}=\{\texttt{FB}_1,\texttt{FB}_2,\ldots \texttt{FB}_A\},$
where the agent can select between $A$ frequency bands. 
The states, on the other hand, corresponding to different Modulation Schemes, i.e.,
$\mathcal{S}=\{\texttt{MS}_1,\texttt{MS}_2,\ldots \texttt{MS}_S\},$
where $S$ represents the number of available modulation schemes in the system.
 Each modulation scheme offers a unique trade-off between data rate and noise tolerance. Lower-order schemes, like BPSK, are more resilient to noise but provide lower data rates, whereas higher-order schemes, such as 1024-QAM or 2048-QAM, offer higher data rates but require better channel conditions. The environmental states represent the channel conditions, and for our study we consider the following 4 environments,
$ \mathcal{E}=\{ \text{Excellent (E)}, \text{Good (G)}, \text{Fair (F)},  \text{Poor (P)} \}.$
The channel conditions are usually not known to the transceiver, but still they can have much influence on the dynamics of the communication system. Moreover, channel conditions are often modelled by Markovian dynamics, i.e., governed by a transition matrix $q(e'|e)$. This is because the stochastic nature of wireless environments, influenced by factors such as fading, interference, and user mobility, inherently introduces dependencies across time steps.
 

The probability of successful transmission depends on several factors, including the channel condition, the chosen modulation scheme, and the selected frequency band \cite{pan2021success,weber2010overview}. We define $P_{\text{success}}(s, e, a)$ as the probability of successful transmission under a given channel condition (environmental state $e$), modulation scheme (system state $s$), and action (frequency band $a$). 
The probability of a successful transmission, $P_{\text{success}}(\cdot)$ for each combination of modulation schemes, selected frequency bands, and channel conditions dictates  the transition probabilities $p_e(s'|s, a)$, as detailed in the Supplementary Materials, see also, e.g.,~\cite{halloush2022formula}.







The reward function $\vec{R}(s, e)$ indicates system performance by considering both data throughput and the cost associated with using higher-order modulation schemes in poor channel conditions. It is defined as:
\begin{align*} 
\vec{R}(s, e) =& \alpha \cdot \text{Rate}(s) \cdot \text{Decay}(e) 
- \beta \cdot \text{Decay}(e)
\end{align*}
where $\alpha$ represents a weight that controls the contribution of the data rate to the overall reward, while $\beta$ serves as a penalty factor for selecting higher-order modulation schemes in suboptimal channel conditions. The term $\text{Rate}(s)$ refers to the data transmission rate associated with a given modulation scheme, and $\text{Decay}(e)$ captures the degradation of system performance based on the current channel condition. Together, these parameters influence the balance between maximizing data throughput and mitigating the risks of poor channel quality. The introduced reward function and state transition probability are only used to make a setting for simulation and are not inferred from the literature.


The agent aims to determine the optimal frequency band for each modulation scheme by taking into account the system's priority of maximizing data throughput while minimizing the impact of channel low quality. The SNS-MDP framework is well-suited for modeling this adaptive communication scenario, where unobservable environment changes occur following a Markov chain. This framework enables algorithms to estimate policy values and apply policy improvement techniques effectively. We illustrate this with a problem involving $S = 11$ modulation schemes and $A = 11$ frequency bands. The detailed model parameters used in the simulations are provided in the Supplementary Material. 
Figure~\ref{fig:VI} illustrates the performance of the TD-learning algorithm for policy evaluation  in Eq~\eqref{Equ:TDlearning-adopted} with a fixed policy, using a constant learning rate of $\alpha = 0.01$ and $\gamma=0.97$. The red curve represents the average performance across 10 independent runs ($M=10$), while the black line indicates the true SNS value $\vec{v}^{\texttt{SNS}}$ as derived in Theorem~\ref{Thm:value_closedform}. The results show that the algorithm converges close to the true value. However, because a fixed learning rate is used rather than a diminishing one as specified in Theorem~\ref{Thm:main}, the algorithm stabilizes within a small region around the fixed point and remains there, which is consistent with the expected behavior of stochastic algorithms with a constant step size.

Figure~\ref{fig:PI-DP} illustrates the performance of the Policy iteration algorithm in Eq.~\eqref{eq:monotonic_improvement_PI_eq_1}. In this experiment,  the agent can evaluate the true SNS Q-table $\vec{Q}^{\texttt{SNS},\mu}$ for a fixed policy. The red curve represents the performance of the Policy iteration algorithm, while the black line indicates the optimal value. The results show that the Policy Improvement algorithm converges to the optimal policy in only a few iterations, thus establishing its efficiency and effectiveness in rapidly finding the optimal solution. This affirms the results in Theorem \ref{Thm:monotonic_improvement} and Theorem \ref{Thm:monotonic_improvement_PI} that establish the convergence and optimally of the Policy Improvement in SNS-MDPs.
Figure~\ref{fig:QL} demonstrates the performance of the Q-learning algorithm. To compute the optimal Q-function, we first determine the optimal value function using policy iteration and then apply Eq.~\eqref{eq:QFunc_ValueFunc}. The red curve represents the Euclidean distance between the Q-function estimated by the proposed algorithm and the derived optimal Q-function, thereby confirming the convergence results established in Theorem \ref{Thm:QL_convergence} within the SNS-MDP framework.

\begin{figure}[t!]
    \centering
    \begin{subfigure}[t]{0.32\textwidth}
        \includegraphics[width=\linewidth]{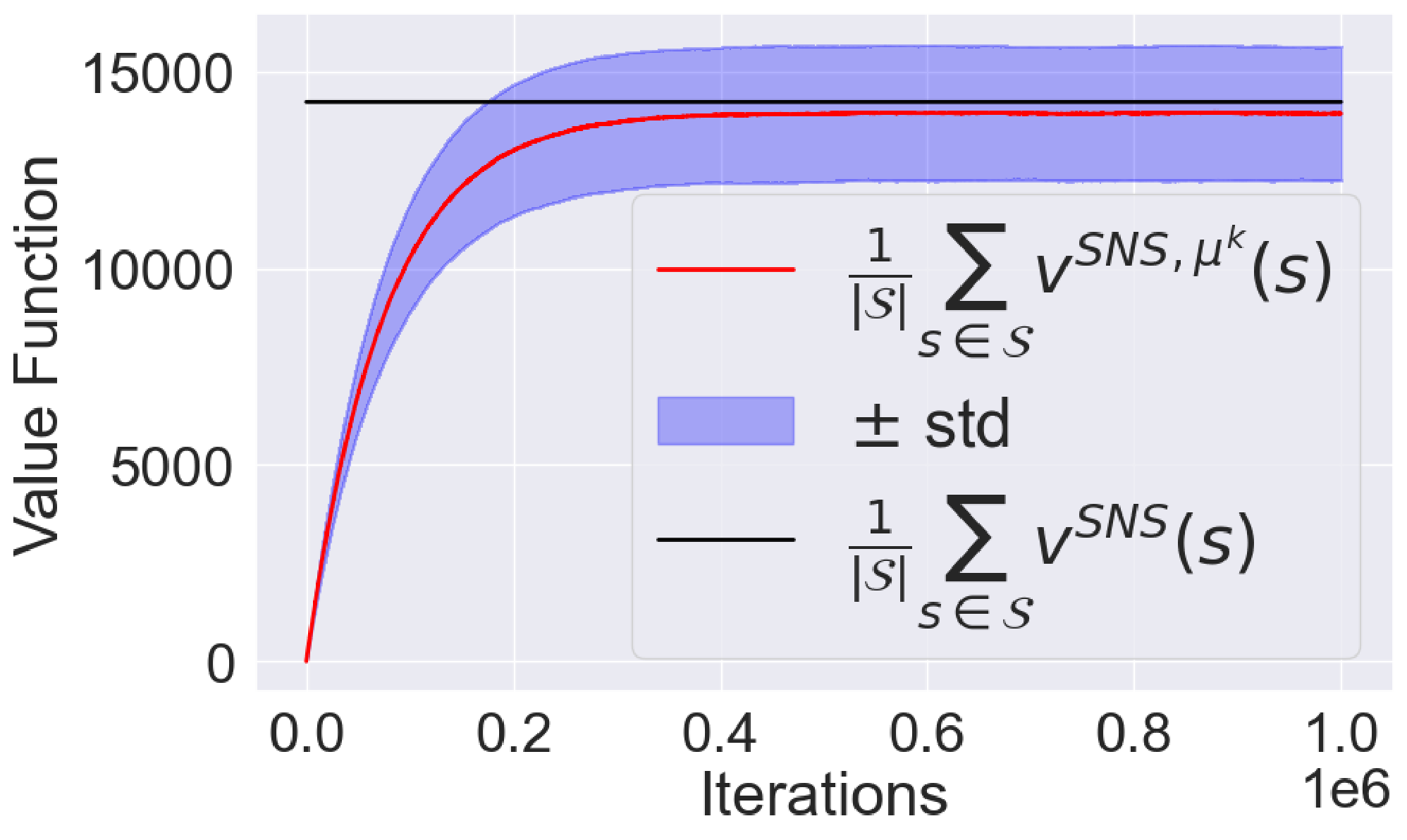} 
        \caption{Policy Evaluation}
        \label{fig:VI}
    \end{subfigure}\hfill
    \begin{subfigure}[t]{0.32\textwidth}
        \includegraphics[width=\linewidth]{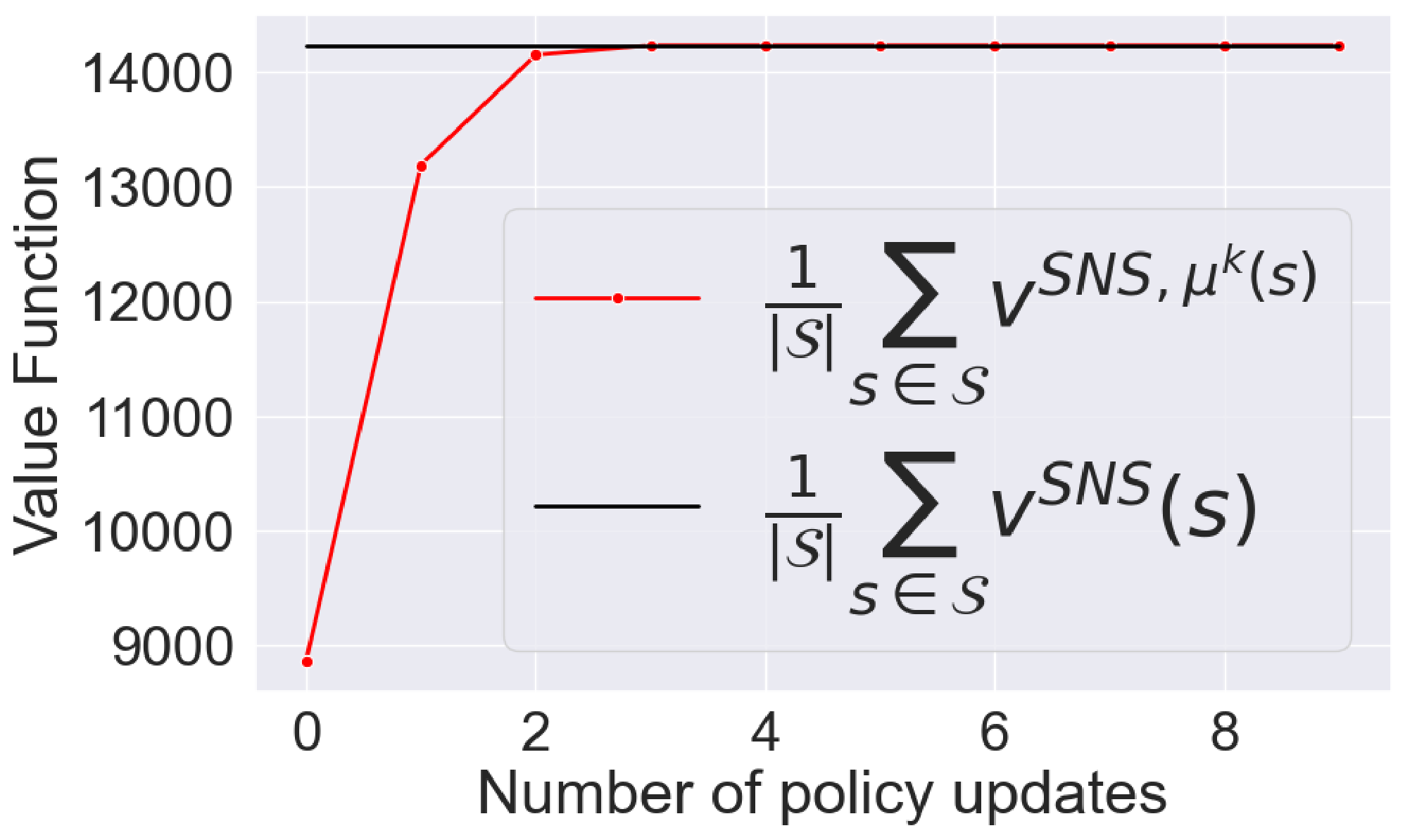} 
        \caption{Policy Iteration}
        \label{fig:PI-DP}
    \end{subfigure}\hfill
    \begin{subfigure}[t]{0.32\textwidth}
        \includegraphics[width=\linewidth]{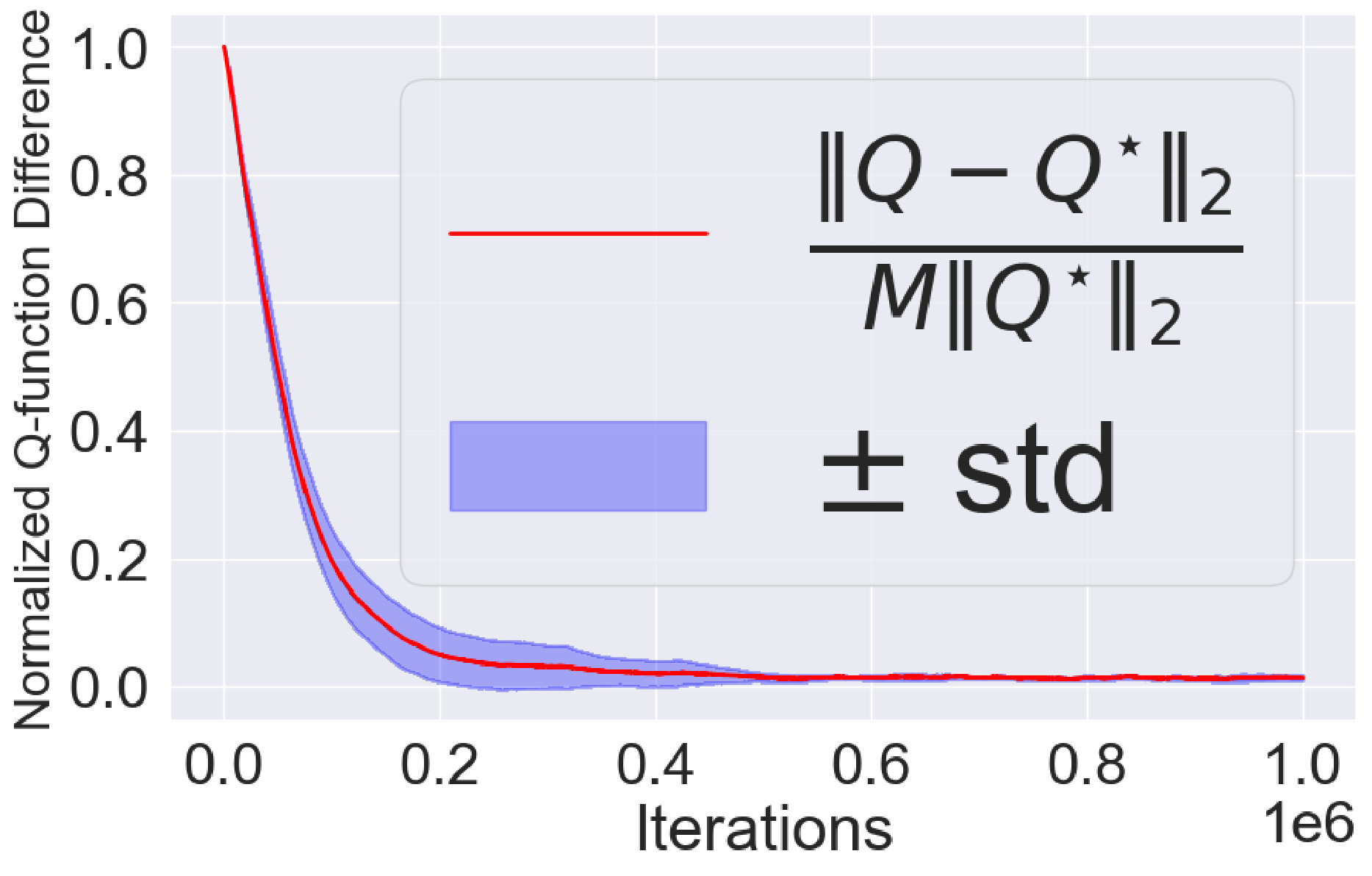} 
        \caption{Q-learning ($M=10$)}
        \label{fig:QL}
    \end{subfigure}
    
    \caption{Convergence of the value iteration, policy iteration, and Q-learning.}
    \label{fig:unified}
\end{figure}

\section{Conclusion}

In this paper, we introduced the SNS-MDP, a novel framework for modeling non-stationary environments driven by an underlying Markov chain. We defined an SNS value function for fixed policies or MRPs and derived a closed-form expression explicitly linked to the Markov chain's statistics. We proved the almost sure convergence of TD-learning algorithms to the SNS value function under fixed policies, despite environmental non-stationarity. Furthermore, we demonstrated policy improvement feasibility and proved the convergence of the policy iteration algorithm toward optimal policies. Additionally, we established the almost sure convergence of Q-learning to an optimal SNS-MDP Q-function under a fixed behavioral policy. The practicality of the framework was validated through application to communication network problems with Markovian channel noise. Future work includes examining additional on-policy and off-policy algorithms, applying the SNS-MDP to multi-task learning, and extending it to multi-agent reinforcement learning.

\section{Acknowledgment}

This work was partially supported by  the Sweden’s Innovation Agency
(Vinnova) and the Swedish Research Council (VR).

\bibliography{refs}
\bibliographystyle{splncs04}

\newpage

\setcounter{equation}{17}
\setcounter{lemma}{2}

\section{Proof of Theorem 1}



The following lemma will be useful for proving the theorem.  

%

\begin{lemma} \label{Lemma:h}
     The sequence $Y_k=(S_k, E_k)$ is a Markov chain $(\mathcal{Y},h(\cdot))$ where the state space is
     \begin{align*} 
\mathcal{Y} = \{(s,e) \in \mathcal{S} \times \mathcal{E} \}
\end{align*}
    and  the transition from state \( y = (s, e) \in \mathcal{Y} \) to state \( y' = (s', e') \in \mathcal{Y} \) is defined by:
\begin{align}\label{eq:h}
h(y'|y) = p_e(s'|s)q(e'|e).
\end{align}

 \end{lemma}
\begin{proof} To prove the lemma, first note that for all $k\in \mathbb{N}$ and $y_i=(s_i,e_i)\in \mathcal{Y}$, for $i=0,\ldots,k$, we have by the chain rule of probability that
\begin{align*} 
&\Pr{Y_k=y_k \mid Y_{k-1}=y_k, \ldots , Y_{0}=y_0} \\ 
&= \Pr{S_k=s_k,E_k=e_k \mid S_{k-1}=s_{k-1},E_{k-1}=e_{k-1},\ldots, S_0=s_0,E_0=e_0}\\
&= \tilde{P}_1 \tilde{P}_2
\end{align*}
where
\begin{align*} 
&\tilde{P}_1 = \Pr{S_k=s_{k} \mid E_k=e_{k}, S_{k-1}=s_{k-1}, E_{k-1}=e_{k-1}, \ldots , S_{0}=s_{0}, E_{0}=e_{0}} \\
&\tilde{P}_2 = \Pr{E_k=e^{k} \mid S_{k-1}=s_{k-1}, E_{k-1}=e_{k-1}, \ldots, S_{0}=s_{0}, E_{0}=e_{0}}.
\end{align*}
By Definition 1, $S_k$ depends only on $S_{k-1}$ and $E_{k-1}$, which yields 
\begin{align}\label{eq:tildeP1}
 \tilde{P}_1=\Pr{S_k=s_{k} \mid  S_{k-1}=s_{k-1}, E_{k-1}=e_{k-1}} = p_{e^{k-1}}(s_k\mid s_{k-1}).
\end{align}
Similarly, $E_k$ depends only on $E_{k-1}$, which yields 
\begin{align}\label{eq:tildeP2}
\tilde{P}_2 = \Pr{E_k=e^{k} \mid E_{k-1}=e_{k-1}} = q(e_k|e_{k-1}).
\end{align}
Therefore, $Y_k$ is a Markov chain. Moreover, by Eq.~\eqref{eq:tildeP1} and~\eqref{eq:tildeP2} we have for any $ y'= (s', e')\in \mathcal{Y}$ $ y= (s, e)\in \mathcal{Y}$ that
$$h(y'|y)=\Pr{Y_k=y' \mid Y_{k-1}=y} = \tilde{P}_1  \tilde{P}_2 = p_e(s'\mid s) q(e'|e).$$
\end{proof}

To prove Theorem 1, we first show that
\begin{align} \label{eq:bellman_new_vectorwise}
\vec{v}^{\texttt{SNS}} = \vec{r}_{\mathcal{E}} + \gamma \left( \sum_{e \in \mathcal{E}} \boldsymbol{\pi}_{\mathcal{E}}(e) \vec{P}_e \right) \vec{v}^{\texttt{SNS}}.
\end{align}
To that end, recall the definition 
\begin{align} \label{eq:bellman_new_elementwise}
&\vec{v}^{\texttt{SNS}}(s) = \mathbb{E}_{E \sim \vecpi_{\mathcal{E}}(\cdot)} \left[ \vec{v}(s, E) \right] = \sum_{e \in \mathcal{E}}\vec{v}(s, e)\vecpi_{\mathcal{E}}(e), 
\end{align}
where $\vecpi_{\mathcal{E}}(e)$  is the stationary distribution of the Markov chain $(\mathcal{E},q(\cdot))$, which exists by Assumption 1 (see discussion in Section~\ref{Section:MarkovChains} below).
\clearpage

 To expand the expression in Eq.~\eqref{eq:bellman_new_elementwise} we note that by Lemma~\ref{Lemma:h} we have that
\begin{align*} 
\vec{v}(s, e) =& \mathbb{E} \left[\sum_{k=0}^{\infty} \gamma^k R_k \mid S_0 
= s, E_0 = e \right] \\
=& \mathbb{E} \left[ R_0 + \gamma \sum_{k=1}^{\infty} \gamma^{k-1} R_k \mid S_0 = s, E_0 = e \right] \\
=& \vec{R}(s,e) \\
+&\gamma  \sum_{s' \in \mathcal{S}} \sum_{e' \in \mathcal{E} }   \mathbb{E} \left[ \sum_{k=1}^{\infty} \gamma^{k-1} R_k \mid S_1 = s', E_1 = e' \right]   \Pr{S_1=s', E_1=e' \mid S_0=s, E_0=e} \\
=& \vec{R}(s,e) + \gamma 
 \sum_{s' \in \mathcal{S}} \sum_{e' \in \mathcal{E} }  \vec{v}(s', e')   \Pr{S_1=s', E_1=e' \mid S_0=s, E_0=e} \\
=& \vec{R}(s,e) + \gamma  \sum_{s' \in \mathcal{S}} \sum_{e' \in \mathcal{E} }   \vec{v}(s', e') 
 \Pr{S_1=s' \mid S_0=s, E_0=e} \Pr{E_1=e' \mid  E_0=e} \\
\end{align*}
where in the last equations we utilized the Markov chain property in Lemma \ref{Lemma:h} and the structure of the transition function in Eq.~\eqref{eq:h}. 

Therefore, by expanding Eq.~\eqref{eq:bellman_new_elementwise} we get that
\begin{align} \label{eq:vSNSeq2}
    \vec{v}^{\texttt{SNS}}(s) =& \sum_{e\in \mathcal{E}}  \vec{R}(s,e) \vecpi_{\mathcal{E}}(e)  + \gamma P' = \vec{r}_{\mathcal{E}}(s) + \gamma P'.
\end{align}
where we recall the definition $\vec{r}_{\mathcal{E}}=\vec{R}\vecpi_{\mathcal{E}}$ and have defined
\begin{align} \label{eq:AA}
&P'=\sum_{e \in \mathcal{E}}  \sum_{s' \in \mathcal{S}} \sum_{e' \in \mathcal{E} }   \vec{v}(s', e') \Pr{S_1=s' \mid S_0=s, E_0=e} \Pr{E_1=e' \mid E_0=e}\Pr{E_0=e}.
\end{align}
We can further manipulate $P'$ to express it in a more favorable form. To do that, note that 
\begin{align}
    \Pr{E_1=e' \mid E_0=e} =& \Pr{E_1=e' \mid S_0=s, E_0=e} \label{eq:1A}\\
    \Pr{E_0=e} =& \Pr{E_0=e \mid S_0=s} \label{eq:1B} \\
    \Pr{S_1=s' \mid S_0=s, E_0=e} =& \Pr{S_1=s' \mid E_1=e', S_0=s, E_0=e}, \label{eq:1C}
\end{align}
where we obtain Eq.~\eqref{eq:1A} and Eq.~\eqref{eq:1B} by the fact that $E_0$ and $E_1$ do not depend on $S_0$ and we obtain Eq.~\eqref{eq:1B} by the fact that $S_1$ only depends on $E_0$ and $S_0$ and not on $E_1$. Moreover, by using the chain rule, we have
\begin{align} \label{eq:1D}
    &\Pr{S_1=s', E_1=e', E_0=e \mid S_0=s} =  \Pr{S_1=s'\mid S_0=s, E_0=e} \\
    & \Pr{E_1=e' \mid S_0=s, E_0=e}\Pr{E_0=e \mid S_0=s}.
\end{align}
By applying first Eq.~\eqref{eq:1A}-\eqref{eq:1C} and then Eq.~\eqref{eq:1D} in Eq~\eqref{eq:AA} we get that
\begin{align} 
P'&=\sum_{s' \in \mathcal{S} } \sum_{e' \in \mathcal{E} }\vec{v}(s', e') \sum_{e \in \mathcal{E}} \Pr{S_1=s' \mid S_0=s, E_0=e}\Pr{E_1=e' \mid S_0=s, E_0=e}\Pr{E_0=e \mid S_0=s}  \notag \\
& = \sum_{s' \in \mathcal{S} } \sum_{e' \in \mathcal{E} } \vec{v}(s', e') \sum_{e \in \mathcal{E}} \Pr{S_1=s', E_1=e', E_0=e \mid S_0=s} \notag \\
& = \sum_{s' \in \mathcal{S} } \sum_{e' \in \mathcal{E} } \vec{v}(s', e') \Pr{S_1=s', E_1=e' \mid S_0=s} \notag \\
& = \sum_{s' \in \mathcal{S} } \sum_{e' \in \mathcal{E} } \vec{v}(s', e') \Pr{E_1=e' \mid S_0=s}\Pr{S_1=s' \mid E_1=e', S_0=s} , \label{eq:1E}
\end{align}
where the third equation is obtained by the fact that the inner most sum is over all $e\in\mathcal{E}$ and the final equation is  obtained by using the chain rule. By noting that $S_1$ does not depend on $E_1$, it only depends on $E_0$, we further get
$$ \Pr{S_1=s' \mid E_1=e', S_0=s}=  \Pr{S_1=s' \mid S_0=s} $$
which allows us to reduce~\eqref{eq:1E} to the following form (after rearranging the terms) 
\begin{align*} 
P' =& \sum_{s' \in \mathcal{S}} \Pr{S_1=s' \mid S_0=s}  \left( \sum_{ e' \in \mathcal{E} } \vec{v}(s', e')\Pr{E_1=e'} \right). 
\end{align*}
 Note that since $E_0\sim \vecpi_{\mathcal{E}}(\cdot)$, where $\vecpi_{\mathcal{E}}(e)$  is the stationary distribution, and because the stationary distribution is invariant under the transition dynamics, we also have that $E_1\sim \vecpi_{\mathcal{E}}(\cdot)$. This means that 
\begin{align} 
P' =& \sum_{s' \in \mathcal{S}} \Pr{S_1=s' \mid S_0=s}  \left( \sum_{ e' \in \mathcal{E} } \vec{v}(s', e')\vecpi_{\mathcal{E}}(e')\right) \notag \\
=& \sum_{s' \in \mathcal{S}} \Pr{S_1=s' \mid S_0=s} \vec{v}^{\texttt{SNS}}(s'), \label{eq:2A}
\end{align}
where we have used the definition of $\vec{v}^{\texttt{SNS}}(s)$ in Eq.~\eqref{eq:bellman_new_elementwise} to obtain the second equality. Moreover,  we have that
\begin{align*} 
\Pr{S_1=s' \mid S_0=s} =& \sum_{e \in \mathcal{E}} \Pr{S_1=s', E_0=e \mid S_0=s}  \\
 =& \sum_{e \in \mathcal{E}} \Pr{S_1=s' \mid E_0=e, S_0=s}\Pr{E_0=e} \\
  =& \sum_{e \in \mathcal{E}} p_e(s'|s) 
\boldsymbol{\pi}_{\mathcal{E}}(e) = \sum_{e \in \mathcal{E}} \boldsymbol{\pi}_{\mathcal{E}}(e) \vec{P}_e(s,s'), 
\end{align*}
where we have used the chain rule in the second equality and the definition of the transition matrix $\vec{P}_e$ in the final equality. Plugging this into Eq.~\eqref{eq:2A} we get
\begin{align*}
    P'= \sum_{s' \in \mathcal{S}}  \sum_{e \in \mathcal{E}} \boldsymbol{\pi}_{\mathcal{E}}(e) \vec{P}_e(s,s')\vec{v}^{\texttt{SNS}}(s') =
   \sum_{e \in \mathcal{E}}     \boldsymbol{\pi}_{\mathcal{E}}(e)  \sum_{s' \in \mathcal{S}} \vec{P}_e(s,s')\vec{v}^{\texttt{SNS}}(s') .
\end{align*}
It is easily checked that this is entry $s$ in the matrix 
$$\sum_{e \in \mathcal{E}}     \boldsymbol{\pi}_{\mathcal{E}}(e)  (\vec{P}_e \vec{v}^{\texttt{SNS}})= \left( \sum_{e \in \mathcal{E}}     \boldsymbol{\pi}_{\mathcal{E}}(e)  \vec{P}_e \right) \vec{v}^{\texttt{SNS}} .$$
Now going back to Eq.~\eqref{eq:vSNSeq2}, we get that
\begin{align*} 
    \vec{v}^{\texttt{SNS}}(s) =&  \vec{r}_{\mathcal{E}}(s) + \gamma P'   \\
    =&  \vec{r}_{\mathcal{E}}(s) + \gamma \sum_{e \in \mathcal{E}}     \boldsymbol{\pi}_{\mathcal{E}}(e)  \sum_{s' \in \mathcal{S}} \vec{P}_e(s,s')\vec{v}^{\texttt{SNS}}(s')
\end{align*}
or in matrix form we get the desired result that
\begin{align}  \label{eq:vSNS_linear_system}
\vec{v}^{\texttt{SNS}} = \vec{r}_{\mathcal{E}} + \gamma \left( \sum_{e \in \mathcal{E}} \boldsymbol{\pi}_{\mathcal{E}}(e) \vec{P}_e \right) \vec{v}^{\texttt{SNS}}.
\end{align}
Since 
$$ \sum_{e \in \mathcal{E}} \boldsymbol{\pi}_{\mathcal{E}}(e) \vec{P}_e$$
is a convex combination of stochastic matrices and $\gamma\in[0,1)$ we know that 
$$\vec{I}-  \gamma  \sum_{e \in \mathcal{E}} \boldsymbol{\pi}_{\mathcal{E}}(e) \vec{P}_e $$
is non-singular matrix and thus the linear system in Eq.~\eqref{eq:vSNS_linear_system} has the unique solution
$$ \vec{v}^{\texttt{SNS}} =
\left(\vec{I} - \gamma\left( \sum_{e \in \mathcal{E}} \boldsymbol{\pi}_{\mathcal{E}}(e) \vec{P}_e \right) \right)^{-1} \vec{r}_{\mathcal{E}}
.$$

\section{Proof of Theorem 2}

To prove Theorem 2 we draw on the following classic result for stochastic systems, see, e.g., Proposition 4.8 in \cite{bertsekas1996neuro}.

\begin{proposition}\label{Prop:Bertsekas}
Consider a finite state Markov chain \((\mathcal{X},z(\cdot))\) with a finite state space \(\mathcal{X}\) and a state sequence:
\begin{equation}\label{eq:trajectory_X}
    X_0,X_1,\ldots,X_k,\ldots ~.
\end{equation}
Let the functions \(\vec{A}:\mathcal{X}\rightarrow \mathbb{R}^{n\times n}\) and \(\vec{b}:\mathcal{X}\rightarrow \mathbb{R}^{n}\) govern an algorithm that generates the sequence \(\vec{v}_k\in \mathbb{R}^n\) according to:
\begin{align} \label{eq:general_algorithm}
    \vec{v}_{k+1}=\vec{v}_k+\alpha(\vec{A}(X_k)\vec{v}_k+\vec{b}(X_k)), 
\end{align}
where \(\alpha_k>0\) is the step-size and \(\vec{v}_0\in \mathbb{R}^n\) is the initialization. Assume the following conditions are met:
\begin{enumerate}
    \item \label{Condition:C1} The step sizes \(\alpha_k\) are deterministic and satisfy the condition
    \begin{align*}
    \sum_{k=0}^{\infty} \alpha_k = \infty \quad \text{and} \quad \sum_{k=0}^{\infty} \alpha_k^2 < \infty.
\end{align*}
    \item \label{Condition:C2} The Markov chain \((\mathcal{X},z(\cdot))\) has an invariant distribution denoted by  \(\boldsymbol{\pi}\in [0,1]^{|\mathcal{X}|}\).
    \item  \label{Condition:C3} The matrix \(\vec{A}=\mathbb{E}_{X\sim \boldsymbol{\pi}} [\vec{A}(X)]\) is  negative-definite. 
    \item \label{Condition:C4} There exists $M>0$ such that \(\|\vec{A}(x)\|\leq M\) and \(\|\vec{b}(x)\|\leq M\) for all \(x \in \mathcal{X}\). 
    \item \label{Condition:C5} There exist constants \(D\in\mathbb{R}_+\) and \(\lambda\in [0,1)\) exist such that:
    \begin{align*}
        \left\| \mathbb{E}[ \vec{A}(X_k)|X_0=X]-\vec{A}  \right\| \leq& D \lambda^k, \\
        \left\| \mathbb{E}[ \vec{b}(X_k)|X_0=X]-\vec{b} \right\| \leq& D \lambda^k,
    \end{align*}
    where \(b=\mathbb{E}_{X\sim \boldsymbol{\pi}} [\vec{b}(X)]\) are valid for all \(k \in \mathbb{N}\) and \(X \in \mathcal{X}\).
\end{enumerate}
Under these conditions, the algorithm's iterates converge with probability one to the unique fixed-point:
\[ \lim_{k\rightarrow \infty} \vec{v}_k= -\vec{A}^{-1}\vec{b}.\]
\end{proposition}

To prove Theorem~2, we construct a Markov chain $(\mathcal{X},z(\cdot))$ along with $\vec{A}$ and $\vec{b}$, as in Proposition~\ref{Prop:Bertsekas}, so that the algorithm in Eq.~\eqref{eq:general_algorithm}  is equivalent to the TD-learning algorithm in Eq.~(9). We then verify that all the conditions of Proposition~\ref{Prop:Bertsekas} are satisfied, thereby confirming that the fixed-point is the unique solution to the system, ensuring convergence of the algorithm to the desired value.

 In particular, we let the Markov chain sequence in Eq.~\eqref{eq:trajectory_X} be such that $X_k=(S_k,S_{k+1},E_k)$. The state space is
 \begin{align}\label{eq:Ki_def}
\mathcal{X} = \{(s, s', e) \in \mathcal{S} \times \mathcal{S} \times \mathcal{E} \mid p_{e}(s'|s)  > 0\},
\end{align}
 where the condition $p_{e}(s'|s)  > 0$ is included since we only consider states   $X_k=(S_k,S_{k+1},E_k)$ where transition from $S_k$ to $S_{k+1}$ is possible. This sequence is indeed a Markov chain. 
 \begin{lemma} \label{Lemma:z}
     The sequence $X_k=(S_k,S_{k+1},E_k)$ is a Markov chain $(\mathcal{X},z(\cdot))$ where the transition from state \( x = (s_1, s_2, e) \in \mathcal{X} \) to state \( x' = (s_1', s_2', e') \in \mathcal{X} \) is defined by:
\begin{equation}\label{eq:z}
    z(x'|x) = \begin{cases}
        p_{e'}(s_2'|s_2) q(e'|e) & \text{if } s_2 = s_1', \\
        0 & \text{otherwise.}
    \end{cases}
\end{equation}
 \end{lemma}

\begin{proof}
%
%
%
First consider the case when $s_2 \neq s_1'$. Since $z(x'|x)$ is the transition probability from $X_k=(S_k,S_{k+1},E_k)$ to $X_{k+1}=(S_{k+1},S_{k+2},E_{k+1})$, $s_2 \neq s_1'$ is the event that $S_{k+1}\neq S_{k+1}$  which is is not possible. Therefore, the probability of this event is zero, i.e., $z(x'|x)=0$.

Consider now the case when \( s_2 = s_1' \). By applying the chain rule of probability to $z(x'|x)$, and recalling that $s_2 = s_1'$, we have
\begin{align*}
    z(x'|x) &= \Pr{X_{k} = (s_2, s_2', e') \mid X_{k-1} = (s_1, s_2, e)} \\
    &= \textbf{Pr} \big[ S_{k+1} = s_2',S_{k} = s_2, E_k = e' \mid S_{k} = s_2, S_{k-1} = s_1, E_{k-1} = e \big] \\
  &= P_1\times P_2\times P_3
\end{align*}
where 
\begin{align*}
    P_1 &= \textbf{Pr} \big[ S_{k+1} = s_2' \mid S_k = s_2, E_k = e', S_{k-1} = s_1, E_{k-1} = e \big], \\
    P_2 &= \textbf{Pr} \big[ S_k = s_2 \mid E_k = e', S_k = s_2, S_{k-1} = s_1, E_{k-1} = e \big], \\
    P_3 &= \Pr{E_k = e' \mid S_k = s_2, S_{k-1} = s_1, E_{k-1} = e}.
\end{align*} 
 Regarding $P_1$, note that by our definition of SNS-MRP, $S_{k+1}$ depends only on $S_k$ and $E_k$  via the transition function $p_{e'}(s_2'|s)$, and, in particular, it does not depend on $E_{k-1}$ or $S_{k-1}$. Therefore, we have
 \begin{align*}
     P_1 =&  \Pr{ S_{k+1} = s_2' \mid S_k = s_2, E_k = e', S_{k-1} = s_1, E_{k-1} = e } \\
     =& \Pr{ S_{k+1} = s_2' \mid S_k = s_2, E_k = e' } \\
     =& p_{e'}(s_2'|s_2).
 \end{align*}
Regarding \( P_2 \), it is evident that \( P_2 = 1 \), as the conditional probability of the event \( S_k = s_2 \) given that \( S_k = s_2 \) is clearly one. Finally, regarding $P_3$, by definition of the Markov Chain $(\mathcal{E},q(\cdot))$, $E_k$ depends only on $E_{k-1}$, and thus we have
\begin{align*}
    P_3 =& \Pr{E_k = e' \mid  E_{k-1} = e}= q(e'|e).
\end{align*}
Therefore, by combining the results above, we get that
$$z(x'|x)=P_1\times P_2 \times P_3 = p_{e'}(s_2'|s_2) q(e'|e).$$
\end{proof}

For a given sample $X_k = (S_k, S_{k+1},E_k)$, define:
\begin{align}
	\vec{A}(X_k) &= \gamma \vec{e}_{\mathcal{S}}(S_k) \vec{e}_{\mathcal{S}}(S_{k+1})^{\mathrm{T}} - \vec{e}_{\mathcal{S}}(S_k) \vec{e}_{\mathcal{S}}(S_k)^{\mathrm{T}} \label{eq:Afun}\\
	\vec{b}(X_k) &= \vec{e}_{\mathcal{S}}(S_k)\vec{e}_{\mathcal{S}}(S_K)^{\mathrm{T}} \vec{R} \vec{e}_{\mathcal{E}}(E_k) \label{eq:bfun}
\end{align}
where \(\vec{e}_{\mathcal{S}}(s) \in \mathbb{R}^{|\mathcal{S}|}\) and \(\vec{e}_{\mathcal{E}}(e) \in \mathbb{R}^{|\mathcal{E}|}\) are unit vectors with a 1 in the \(s\)-th and \(e\)-th position. It is easy to verify that with this definition of    \(\vec{A}(\cdot)\) and \(\vec{b}(\cdot)\), the algorithm in Eq.~\eqref{eq:general_algorithm} of Proposition~\ref{Prop:Bertsekas} is equivalent to TD algorithm as described in Eq.~(9).


In subsections~\ref{sec:Proofb} and~\ref{sec:Proofcd}, we establish that $(\mathcal{X}, z(\cdot))$ possesses an invariant distribution $\vecpi$ and confirm that
\begin{align}
    \vec{A} &= \mathbb{E}_{X \sim \vecpi(\cdot)}[\vec{A}(X)] = \gamma \vec{D}_{\boldsymbol{\pi}_{\mathcal{S}}} \left( \sum_{e \in \mathcal{E}} \boldsymbol{\pi}_{\mathcal{E}}(e) \vec{P}_e \right) - \vec{D}_{\boldsymbol{\pi}_{\mathcal{S}}}, \label{eq:A}\\
    \vec{b} &= \mathbb{E}_{X \sim \vecpi(\cdot)}[\vec{b}(X)] = \vec{D}_{\boldsymbol{\pi}_{\mathcal{S}}} \vec{r}_{\mathcal{E}} . \label{eq:b}
\end{align}
where \(\vec{D}_{\boldsymbol{\pi}_{\mathcal{S}}} = \text{Diag}(\boldsymbol{\boldsymbol{\pi}}_{\mathcal{S}})\) and $\vecpi_{\mathcal{S}}$ is defined in subsection~\ref{sec:Proofcd}. Therefore, if we verify that conditions (a)-(e) of Proposition~\ref{Prop:Bertsekas} are satisfied, then the TD-learning algorithm converges with probablity one to the fixed-point
\begin{align*}
    \lim_{k\rightarrow \infty} \vec{v}_k =&-\vec{A}^{-1}\vec{b} =\left(\vec{D}_{\boldsymbol{\pi}_{\mathcal{S}}}\left(\vec{I} - \gamma \left( \sum_{e \in \mathcal{E}} \boldsymbol{\pi}_{\mathcal{E}}(e) \vec{P}_e \right)\right)\right)^{-1} \vec{D}_{\boldsymbol{\pi}_{\mathcal{S}}} \vec{r}_{\mathcal{E}} \\
    =& \left(\vec{I} - \gamma \left( \sum_{e \in \mathcal{E}} \boldsymbol{\pi}_{\mathcal{E}}(e) \vec{P}_e \right)\right)^{-1} \vec{r}_{\mathcal{E}},
\end{align*}
and the proof of Theorem~2 is complete. We note that condition (a) holds trivially; the step-sizes are already assumed to satisfy this condition. The subsequent subsections are devoted to the validation of conditions (b)-(e).

\subsection{Condition \ref{Condition:C2})} \label{sec:Proofb}

We now demonstrate that the Markov chain $(\mathcal{X},z(\cdot))$ possesses an invariant distribution $\boldsymbol{\pi}$. According to Proposition~\ref{Prop:invariantDist} in subsection~\ref{SubSection:P_of_MCs}, it suffices to establish that $(\mathcal{X},z(\cdot))$ is irreducible and aperiodic.

\begin{lemma} \label{lemma:l1}
    Under Assumption 1, the Markov chain $(\mathcal{X},z(\cdot))$ is irreducible and aperiodic. 
\end{lemma}
\begin{proof}

To prove this result, it is useful to first define the set of all feasible trajectories. Let \( x_{t} = (s_{t}, s_{t}', e_{t}) \in \mathcal{X} \) denote the state at time \( t \). However, given the definition of the Markov chain \( (\mathcal{X}, z(\cdot)) \), there is a temporal dependence between the states \( x_t \). In particular, \( s_{t}' \) essentially represents \( s_{t+1} \), meaning that only trajectories where \( s_{t}' = s_{t+1} \) are feasible.
Thus, define the set of all feasible trajectories starting at time \( t = 0 \) and ending at time \( t = k \), with initial value \( x_0 = x^{\texttt{I}} \) and terminal value \( x_k = x^{\texttt{T}} \), as follows:
$$
\mathcal{X}_{k}^{\texttt{Traj}}(x^{\texttt{I}}, x^{\texttt{T}}) = \{ (x_{0}, \ldots, x_{k}) \in \mathcal{X}^{k+1} \mid s_t' = s_{t+1} \text{ for } t = 0, \ldots, k - 1, \, x_0 = x^{\texttt{I}}, \, x_k = x^{\texttt{T}} \}.
$$
We use the following notation for a trajectory
$$\mathbf{x}_{0:k} =(x_{0},\ldots, x_{k})\in\mathcal{X}_{k_1:k_2}^{\texttt{Traj}}$$ 
and to simplify the notation, and since we have the condition $s_t'=s_{t+1}$, we represent a state 
such that $x_{t}
=(s_t,s_{t+1},e_t)$ for $t=k_1,\ldots, k_2$ instead of $x_{t}
=(s_t,s_{t}',e_t)$.

We are now ready to prove the lemma. 
Our proof strategy is to apply Proposition~\ref{Prop:MCequivalence} from subsection~\ref{SubSection:P_of_MCs}. Specifically, by Proposition~\ref{Prop:MCequivalence}, the Markov chain \( (\mathcal{X}, z(\cdot)) \) is irreducible and aperiodic if and only if there exists some \( K \in \mathbb{N} \) such that for all \( x_0, x \in \mathcal{X} \), the following condition holds:
\begin{equation} \label{eq:z_condition_K}
    z^k(x \mid x_0) > 0 \text{ for all } k \geq K.
\end{equation}
In the remainder of the proof, we will establish the existence of such a \( K \).


We start by expressing $z^k(x_k \mid x_0)$ in a more convenient form. To that end, take any $x_{0},x_k \in \mathcal{X}$. We then have that:
    \begin{align}
        z^{k}(x_{k}|x_0)&= \Pr{X_k=x_k|X_0=x_0} \notag \\
        &= \sum_{\mathbf{x}_{0:k}\in \mathcal{X}_{k}^{\texttt{Traj}}(x_0, x_k)} \Pr{X_k=x_k,X_{k-1}=x_{k-1},\ldots,X_1=x_1|X_0=x_0},  \label{eq:zk_sum}
    \end{align}
    where the second equality comes by the fact that we sum over all possible trajectories starting at state $x_0$ and ending at state $x_k$.  By applying the chain rule of probability and the Markov property recursively,  it is easy to establish that 
  \begin{align*}
    \Pr{X_k=x_k,\ldots,X_1=x_1|X_0=x_0}   
     & = \Pr{X_k=x_k|X_{k-1}=x_{k-1}} \cdots \Pr{X_{1}=x_{1}|X_{0}=x_{0}} \\
     & = z(x_{k}|x_{k-1})\cdots z(x_{1}|x_{0}) \\
     & = p_{e_{k}}(s_{k+1}|s_{k})\cdots p_{e_{0}}(s_{1}|s_{0}) q(e_{k}|e_{k-1})\cdots q(e_{1}|e_{0}),
  \end{align*}
where in the final equation we have used the decomposition of $z(\cdot)$ in Lemma~\ref{Lemma:z}.
Therefore, by further expanding Eq.~\eqref{eq:zk_sum} we get
\begin{align}
        z^{k}(x^{k}|x^0)=& 
        \sum_{\mathbf{x}_{0:k}\in \mathcal{X}_{k}^{\texttt{Traj}}(x_0, x_k)} p_{e^{k}}(s^{k+1}|s^{k})\cdots p_{e^{0}}(s^{1}|s^{0}) q(e^{k}|e^{k-1})\cdots q(e^{1}|e^{0}) \\
        =& \sum_{\mathbf{x}_{0:k}\in \mathcal{X}_{k}^{\texttt{Traj}}(x_0, x_k)}  q(e^{k}|e^{k-1})\cdots q(e^{1}|e^{0}) \Gamma(\mathbf{x}_{0:k}) \label{eq:zkxx}
\end{align}    
where 
$$\Gamma(\mathbf{x}_{0:k})=
p_{e^{k}}(s^{k+1}|s^{k})\cdots p_{e^{0}}(s^{1}|s^{0}). 
$$
Note that by the definition of state space $\mathcal{X}$, $p_{e^{i}}(s^{i+1}|s^{i})>0$ for all $i = 0, 1, 2, \ldots, k$. This means that $\Gamma(\mathbf{x}_{0:k})$ is always positive, i.e., $\Gamma(\mathbf{x}_{0:k})>0$. Therefore, to show that   there exists $K$ such that the condition in Eq.~\eqref{eq:z_condition_K} holds for all $x_0,x\in \mathcal{X}$, where $x_0=(s_0,s_0',e_0)$ and $x=(s,s',e)$, it suffices show that   there exists $K$  such that
\begin{equation} \label{eq:q0dddk}
q(e^{k}|e^{k-1})\cdots q(e^{1}|e^{0})>0~~\text{ for all }k\geq K
\end{equation}
for some trajectory 
$$ e_0,e_1,\ldots, e_k$$
where $e_k=e$. Since by Assumption~1, the Markov chain $(\mathcal{E},q(\cdot))$ is  irreducible and aperiodic, by Proposition~\ref{Prop:MCequivalence}  we know that there exists $K$ such that for all $e_0,e\in \mathcal{E}$ it holds that $q^k(e\mid e_0)>0$ for all $k\geq K$. In particular, there exists a trajectory
$$ e_0,e_1,\ldots, e_k$$
where $e_k=e$ such that 
$$q(e^{k}|e^{k-1})\cdots q(e^{1}|e^{0})>0.$$
Since we can do this  for all $e_0,e\in \mathcal{E}$, we have established that~\eqref{eq:q0dddk} holds for this $K$, which in turn, establishes, by Eq.~\eqref{eq:zkxx}, that the condition in Eq.~\eqref{eq:z_condition_K} holds for the same $K$. Thus by Proposition~\ref{Prop:MCequivalence}  we can conclude that $(\mathcal{X},\vec{z}(\cdot))$ is irreducible and aperiodic.
\end{proof}

\subsection{Conditions \ref{Condition:C3}) and \ref{Condition:C4})} \label{sec:Proofcd}

 We start by proving that Equations~\eqref{eq:A} and~\eqref{eq:b} hold true. Note that for states $s,s'\in \mathcal{S}$ then  $\vec{e}_{\mathcal{S}}(s) \vec{e}_{\mathcal{S}}(s)^{\rm{T}}$ is a $n \times n$ matrix  that is everywhere zero except it has $1$ on the diagonal element corresponding to state $s$. Similarly, $\vec{e}_{\mathcal{S}}(s) \vec{e}_{\mathcal{S}}(s')^{\rm{T}}$ is a $n \times n$ matrix that is everywhere zero except it has $1$ on the row and column corresponding, respectively, to the states $s$ and $s'$. 
 
 Consider the tuple \(x = (s, s', e) \in \mathcal{X}\). 
As established in subsection~\ref{sec:Proofb}, the Markov chain \( (\mathcal{X}, z(\cdot)) \) has a stationary distribution \( \boldsymbol{\pi}(x) = \Pr{s, s', e} \), which represents the probability of being in state \( (s, s', e) \) at equilibrium. It is also useful to define a {marginal stationary distribution} over the state space \( \mathcal{S} \), denoted by \( \boldsymbol{\pi}_{\mathcal{S}} \), which captures the marginal probability of being in state \( s \in \mathcal{S} \) by summing over the remaining variables \( s' \in \mathcal{S} \) and \( e \in \mathcal{E} \). Formally, we define \( \boldsymbol{\pi}_{\mathcal{S}} \) as:
\[
\boldsymbol{\pi}_{\mathcal{S}}(s) = \sum_{e \in \mathcal{E}, s' \in \mathcal{S}} \boldsymbol{\pi}(s, s', e).
\]
 %
 %
 %
%
Additionally, we define the {aggregated state transition matrix}, denoted by \( \boldsymbol{\Pi}_{\mathcal{S}, \mathcal{S}} \), which represents the expected transition dynamics between states in \( \mathcal{S} \) after averaging over the environmental variable \( e \in \mathcal{E} \). This matrix is defined as:
\[
\boldsymbol{\Pi}_{\mathcal{S}, \mathcal{S}} = \sum_{e \in \mathcal{E}} \boldsymbol{\pi}_{\mathcal{E}}(e) \vec{P}_e,
\]
where \( \boldsymbol{\pi}_{\mathcal{E}}(e) \) is, again, the stationary distribution of the environmental Markov chain $(\mathcal{E},q(\cdot))$, and \( \vec{P}_e \) is the transition matrix for a fixed environmental state \( e \).
 It can now be established that:
\begin{align} \label{eq:eeT_identity}
    \mathbb{E}_{X\sim \vecpi(\cdot)} [ \vec{e}_{\mathcal{S}}(s) \vec{e}_{\mathcal{S}}(s)^{\rm{T}} ]=&  \texttt{Diag}(\boldsymbol{\pi}_{\mathcal{S}}) = \boldsymbol{D}_{\boldsymbol{\pi}_{\mathcal{S}}} \\
    \mathbb{E}_{X\sim \vecpi(\cdot)} [ \vec{e}_{\mathcal{S}}(s) \vec{e}_{\mathcal{S}}(s')^{\rm{T}} ]=&  \boldsymbol{D}_{\boldsymbol{\pi}_{\mathcal{S}}} \boldsymbol{\Pi}_{\mathcal{S}, \mathcal{S}}. 
\end{align}

Equation~\eqref{eq:A} now follows directly from the definition of $\vec{A}(\cdot)$ in Equation~\eqref{eq:Afun}. In the same manner, Equation~\eqref{eq:b} is derived by combining Equation~\eqref{eq:eeT_identity} and the definition in Equation~\eqref{eq:bfun}, considering the independence between the current state and the current environmental state. According to the definition of \(\mathcal{X}\), the current environmental state influences the next state, not the current state. Additionally, $\mathbb{E}_{X \sim \vecpi(\cdot)} [\vec{e}_{\mathcal{E}}(e)]= \boldsymbol{\pi}_{\mathcal{E}}$.

We next prove that $\vec{A}$ is negative definite.  To that end, we show that $\vec{w}^{\rm{T}}\vec{A}\vec{w}<0$ for all $\vec{w}\in \R^{|\mathcal{S}|}\setminus \{0\}$. 
In particular, we have that
\begin{align} \notag
    \vec{w}^{\rm{T}}\vec{A}\vec{w}=& \vec{w}^{\rm{T}} \left(  \gamma \boldsymbol{D}_{\boldsymbol{\pi}_{\mathcal{S}}} \boldsymbol{\Pi}_{\mathcal{S}, \mathcal{S}} - \boldsymbol{D}_{\boldsymbol{\pi}_{\mathcal{S}}} \right) \vec{w} \\
       =& \gamma \vec{w}^{\rm{T}} \boldsymbol{D}_{\boldsymbol{\pi}_{\mathcal{S}}} \boldsymbol{\Pi}_{\mathcal{S}, \mathcal{S}} \vec{w} -   \vec{w}^{\rm{T}}  \boldsymbol{D}_{\boldsymbol{\pi}_{\mathcal{S}}} \vec{w}. \label{eq:wAw}
\end{align}
 Let $\boldsymbol{D}_{\boldsymbol{\pi}_{\mathcal{S}}}^{1/2}\in \R^{|\mathcal{S}|\times |\mathcal{S}|}$ be the diagonal matrix whose entries are the element-wise square roots of the corresponding elements in $\boldsymbol{D}_{\boldsymbol{\pi}_{\mathcal{S}}}$.
Then, by the Cauchy–Schwarz inequality, we obtain
\begin{align} \notag
   \vec{w}^{\rm{T}} \boldsymbol{D}_{\boldsymbol{\pi}_{\mathcal{S}}}\boldsymbol{\Pi}_{\mathcal{S}, \mathcal{S}} \vec{w}
        &= \left( \boldsymbol{D}_{\boldsymbol{\pi}_{\mathcal{S}}}^{1/2} \vec{w} \right)^{\rm{T}} \boldsymbol{D}_{\boldsymbol{\pi}_{\mathcal{S}}}^{1/2}\boldsymbol{\Pi}_{\mathcal{S}, \mathcal{S}} \vec{w}  \\
        &\leq ||\boldsymbol{D}_{\boldsymbol{\pi}_{\mathcal{S}}}^{1/2} \vec{w}||_2  ||\boldsymbol{D}_{\boldsymbol{\pi}_{\mathcal{S}}}^{1/2}\boldsymbol{\Pi}_{\mathcal{S}, \mathcal{S}} \vec{w} ||_2. \label{eq:wDPw}
\end{align}
By considering the norm
$$||\vec{w}||_{\boldsymbol{D}_{\boldsymbol{\pi}_{\mathcal{S}}}}=\sqrt{\vec{w}^{\rm{T}}\boldsymbol{D}_{\pi_{\mathcal{S}}}\vec{w}}$$
and using that $ ||\boldsymbol{D}_{\boldsymbol{\pi}_{\mathcal{S}}}^{1/2} \vec{w}||_2=||\vec{w}||_{\boldsymbol{D}_{\pi_{\mathcal{S}}}}$ for all $\vec{w}$ we have that
$$ \vec{w}^{\rm{T}} \boldsymbol{D}_{\boldsymbol{\pi}_{\mathcal{S}}} \boldsymbol{\Pi}_{\mathcal{S}, \mathcal{S}} \vec{w} \leq ||\vec{w}||_{\boldsymbol{D}_{\boldsymbol{\pi}_{\mathcal{S}}}} ||\Pi_{\mathcal{S}, \mathcal{S}}\vec{w}||_{\boldsymbol{D}_{\boldsymbol{\pi}_{\mathcal{S}}}}. $$
It is easily verified that $ ||\boldsymbol{\Pi}_{\mathcal{S}, \mathcal{S}}\vec{w}||_{\boldsymbol{D}_{\boldsymbol{\pi}_{\mathcal{S}}}}\leq ||\vec{w}||_{\boldsymbol{D}_{\boldsymbol{\pi}_{\mathcal{S}}}}$ for all $\vec{w}\in \R^{|\mathcal{S}|}$, see, e.g., Lemma 7.1 in~\cite{tsitsiklis1997analysis}. This, together with Equations~\eqref{eq:wAw} and~\eqref{eq:wDPw} ensures that
\begin{align*}
    \vec{w}^{\rm{T}}\vec{A}\vec{w}\leq \gamma ||\vec{w}||_{\boldsymbol{D}_{\boldsymbol{\pi}_{\mathcal{S}}}}^2 - ||\vec{w}||_{\boldsymbol{D}_{\boldsymbol{\pi}_{\mathcal{S}}}}^2  = (\gamma-1) ||\vec{w}||_{\boldsymbol{D}_{\boldsymbol{\pi}_{\mathcal{S}}}}^2.
\end{align*}
Since $\gamma<1$, it follows that $\vec{w}^{\rm{T}}\vec{A}\vec{w}<0$ for all $\vec{w}\in \R^{|\mathcal{S}|}$.

Finally, we establish that there exists $M\in \R$ such that $||\vec{A}||\leq M$ and $||\vec{b}||\leq M$. To that end, note that the state space $\mathcal{X}$ is finite, thus $\vec{A}(x)$ and $\vec{b}(x)$ can only take finite values, and must thus be bounded for all $x\in \mathcal{X}$, i.e., there exists $M\in \R$ such that $\vec{A}(x)\leq M$ for all $x\in \mathcal{X}$. This means that $||\vec{A}||=||\mathbb{E}_{X\sim \boldsymbol{\pi}} [\vec{A}(X)]||\leq M$  and $||\vec{b}||=||\mathbb{E}_{X\sim \boldsymbol{\pi}} [\vec{b}(X)]||\leq M$, so $\vec{A}$ and $\vec{b}$ are bounded.

\subsection{Condition \ref{Condition:C5})} \label{sec:Proofe}
From Lemma~\ref{lemma:l1} proved above, the Markov chain $(\mathcal{X},z(\cdot))$ is both irreducible and aperiodic. Therefore, by the Convergence Theorem for Markov chains, see, e.g., Theorem 4.9 in Chapter 4 in~\cite{levin2017markov}, there exist $\lambda \in (0,1)$ and $D > 0$ such that for all $x\in \mathcal{X}$ we have
\begin{align*} 
	&\max_{x \in \mathcal{X}} ||z^{k}(\cdot|x)-\vecpi||_{\rm{TV}} \leq D\lambda^{k} \quad \text{ for all} \quad n \in \mathbb{N}.
\end{align*} 
Therefore, recalling from above that there exists $M\in \R$ such that $||A(x)||\leq M$ for all $x\in \mathcal{X}$, we have
\begin{align*} 
	&\|\mathbb{E}[\vec{A}(X_{k})|X_{0}{=}x_{0}]{-}\vec{A} \| = \left\|\sum_{x \in \mathcal{X}}\vec{A}(x)(z^k(x|x_{0}){-}\vecpi(x)) \right\| \\
 &\leq \sum_{x \in \mathcal{X}}\|\vec{A}(x) \| |z^{k}(x|x_{0})-\vecpi(x)| \\ 
&\leq  M \sum_{x \in \mathcal{X}} |z^{k}(x|x_{0})-\vecpi(x)| = 2 M ||z^{k}(\cdot|x_0)-\vecpi||_{\rm{TV}}  \\
 &\leq 2 M D \lambda^k.
\end{align*}
Therefore,  the first inequality in part \ref{Condition:C5}) of Proposition~\ref{Prop:Bertsekas} is established. The second inequality follows similarly
\begin{align*} 
	\|\mathbb{E}[\vec{b}(X_{k})|X_{0}=x_{0}]-\vec{b} \| 
 \leq& M \sum_{x\in \mathcal{X}} |z^k(x|x_0)-\vecpi(x)| \\
 \leq& 2MD \lambda^k.
\end{align*}
As a result, both inequalities of part \ref{Condition:C5}) are established, which concludes the proof.


\section{Proof of Lemma 1} 
To prove the Lemma, first we use the Eq.~(11),
\begin{align*} 
\vec{Q}^{\texttt{SNS},\mu}(s, a) = 
 \mathbb{E}_{E \sim \pi_{\mathcal{E}}(\cdot)} \left[ \vec{Q}^{\mu}(s, E, a) \right]. 
\end{align*}

Thus, from the definition of \( \boldsymbol{Q}^{\texttt{SNS},\mu}(s, e, a) \) we have

\begin{align*}
\boldsymbol{Q}^{\mu}(s, e, a) &= \mathbb{E} \left[ \sum_{k=0}^{\infty} \gamma^k \boldsymbol{R}_k \, \middle| \, S_0 = s, E_0 = e, A_0 = a \right] \\
&= \boldsymbol{R}(s, e, a) + \gamma \mathbb{E} \left[ \sum_{k=1}^{\infty} \gamma^{k-1} \boldsymbol{R}_k \, \middle| \, S_0 = s, E_0 = e, A_0 = a \right] \\
&= \boldsymbol{R}(s, e, a) + \gamma \sum_{s' \in \mathcal{S}} \sum_{e' \in \mathcal{E}} \vec{v}^{\text{SNS},\mu}(s', e')\Pr{ S_1 = s', E_1 = e' \, \middle| \, S_0 = s, E_0 = e, A_0 = a } .
\end{align*}

Therefore, it is only necessary to substitute $\boldsymbol{Q}^{\mu}(s, e, a)$ into Eq.~(11). To achieve this, note that $E_0 \sim \vecpi_{\mathcal{E}}(\cdot)$, where $\vecpi_{\mathcal{E}}(e)$ represents the stationary distribution. Consequently, we have:
\begin{align*} 
\vec{Q}^{\texttt{SNS},\mu}(s, a) =&
 \mathbb{E}_{E \sim \pi_{\mathcal{E}}(\cdot)} \left[ \vec{Q}^{\mu}(s, E, a) \right]=\sum_{e \in \mathcal{E}}\vec{Q}^{\mu}(s, e, a)\Pr{E_0=e} \\
 =&\sum_{e \in \mathcal{E}}\boldsymbol{R}(s, e, a)\Pr{E_0=e} \\
 +& \gamma \sum_{e \in \mathcal{E}}\sum_{s' \in \mathcal{S}} \sum_{e' \in \mathcal{E}} \vec{v}^{\mu}(s', e')\Pr{ S_1 = s', E_1 = e' \, \middle| \, S_0 = s, E_0 = e, A_0 = a }\Pr{E_0=e} \\
 =&\boldsymbol{r}_{\mathcal{E}}(s, a) \\
 +& \gamma \sum_{e \in \mathcal{E}}\sum_{s' \in \mathcal{S}} \sum_{e' \in \mathcal{E}} \vec{v}^{\mu}(s', e')\Pr{ S_1 = s', E_1 = e' \, \middle| \, S_0 = s, E_0 = e, A_0 = a } \Pr{E_0=e}.
 \end{align*}
Here, $\vec{v}^{\mu}(s', e')$ denotes the value function, as defined in Eq.~(4), under the fixed policy $\mu$. Since the environmental state $E_0$ is independent of both $S_0$ and $A_0$, we can express the following equivalence:
\begin{align*} 
\Pr{E_0=e}=\Pr{E_0=e|S_0=s, A_0=a}.
 \end{align*}
 Thus, we have,
 \begin{align*} 
 \vec{Q}^{\texttt{SNS},\mu}(s, a)&=\boldsymbol{r}_{\mathcal{E}}(s, a) \\
 &+ \sum_{s' \in \mathcal{S}}\sum_{e' \in \mathcal{E}}\sum_{e \in \mathcal{E}} \vec{v}^{\mu}(s',e') \Pr{S_1=s', E_1=e'| S_0=s, A_0=a, E_0=e}\Pr{E_0=e| S_0=s, A_0=a} \\
&=\boldsymbol{r}_{\mathcal{E}}(s, a) + \sum_{s' \in \mathcal{S}}\sum_{e' \in \mathcal{E}} \vec{v}^{\mu}(s',e') \sum_{e \in \mathcal{E}}\Pr{S_1=s', E_1=e', E_0=e| S_0=s, A_0=a} \\
&=\boldsymbol{r}_{\mathcal{E}}(s, a) + \sum_{s' \in \mathcal{S}}\sum_{e' \in \mathcal{E}} \vec{v}^{ \mu}(s',e') \Pr{S_1=s', E_1=e'| S_0=s, A_0=a}\\
&=\boldsymbol{r}_{\mathcal{E}}(s, a)  \\
&+ \sum_{s' \in \mathcal{S}}\sum_{e' \in \mathcal{E}} \vec{v}^{\mu}(s',e') \Pr{E_1=e'| S_0=s, A_0=a}\Pr{S_1=s'| S_0=s, A_0=a, E_1=e'} \\
&\quad \text{(Using chain rule)}\\
&=\boldsymbol{r}_{\mathcal{E}}(s, a)  \\
&+ \sum_{s' \in \mathcal{S}} \left(\sum_{e'\in \mathcal{E}} \vec{v}^{\mu}(s',e') \Pr{E_1=e'} \right)\Pr{S_1=s'| S_0=s, A_0=a} \\
&\quad \text{(Using SNS-MDP property)}\\
&=\boldsymbol{r}_{\mathcal{E}}(s, a)  + \sum_{s' \in \mathcal{S}} \vec{v}^{\texttt{SNS}, \mu}(s')  \Pr{S_1=s'| S_0=s, A_0=a} \quad \text{(Using Eq.~(5))} \\
&=\boldsymbol{r}_{\mathcal{E}}(s, a) + \sum_{s' \in \mathcal{S}} \vec{v}^{\texttt{SNS}, \mu}(s')  p(s'| s, a).
\end{align*}

In the equation above, the transition probability $p(s'| s, a)$ exists and can be derived as follows:
\begin{align*} 
p(s'| s, a)&=\sum_{e \in \mathcal{E}} \pi_{\mathcal{E}}(e)p_e(s'| s, a).
\end{align*}

\section{Proof of Theorem 3} 

To prove this Theorem, we utilize Lemma 1 to compute $\boldsymbol{Q}^{\texttt{SNS}, \mu}(s, \mu')$ as follows:
\begin{align} \label{eq:QFunc_new_policy}
\boldsymbol{Q}^{\texttt{SNS}, \mu}(s, \mu') =& \mathbb{E}_{A_0 \sim \mu'(.|s)}[\boldsymbol{Q}^{\texttt{SNS}, \mu}(s, A_0)] \\
 =& 
\mathbb{E}_{\mu'} \left[ \boldsymbol{r}_{\mathcal{E}}(S_0, A_0) +\gamma \vec{v}^{\texttt{SNS}, \mu}(S_1) |S_0=s    \right],
%
\end{align}
where $\mathbb{E}_{\mu'}$ denotes the expected value when we follow the policy $\mu'$.
Note that by the assumption of the theorem, $\forall s \in \mathcal{S}$ we have,
\begin{align} \label{eq:QFunc_new_policy_geq_value}
\vec{v}^{\texttt{SNS}, \mu}(s) \leq \boldsymbol{Q}^{\texttt{SNS}, \mu}(s, \mu') .
\end{align}
We can now derive the result by recursively applying Eq.~\eqref{eq:QFunc_new_policy} and Eq.~\eqref{eq:QFunc_new_policy_geq_value} as follows
\begin{align*}
   \vec{v}^{\texttt{SNS}, \mu}(s)  
  \leq& \vec{Q}^{\texttt{SNS},\mu} (s,\mu') \\
  =&  \mathbb{E}_{\mu'} \left[ \vec{r}_{\mathcal{E}}(S_0, A_0) + \gamma \vec{v}^{\texttt{SNS},\mu}(S_1) \mid S_0=s \right] ~~~~~~~~~~~~~~~~~~~~~~~~\text{(Using \eqref{eq:QFunc_new_policy})} \\
  \leq&  \mathbb{E}_{\mu'} \left[ \vec{r}_{\mathcal{E}}(S_0, A_0) + \gamma \vec{Q}^{\texttt{SNS},\mu}(S_1,\mu') \mid S_0=s \right] ~~~~~~~~~~~~~~~~~~~~\text{(Using \eqref{eq:QFunc_new_policy_geq_value})}  \\
  =& \mathbb{E}_{\mu'} \left[ \vec{r}_{\mathcal{E}}(S_0, A_0) + \gamma \vec{r}_{\mathcal{E}}(S_1, A_1) + \gamma^2 \vec{v}^{\texttt{SNS},\mu}(S_2) \mid S_0=s \right] ~~~~~~~~~\text{(Using \eqref{eq:QFunc_new_policy})} \\
  \leq& ~~~\cdots \\
  =& \mathbb{E}_{\mu'} \left[ \sum_{k=0}^{\infty} \gamma^k \vec{r}_{\mathcal{E}}(S_k, A_k) \mid S_0=s \right] = \vec{v}^{\texttt{SNS}, \mu'}(s)
\end{align*}

\section{Proof of Theorem 4} 

Since $\mu^{n+1}=\mu^n$ we also have that \(\vec{v}^{\mu^{n+1}}(s) = \vec{v}^{\mu^n}(s)\) for all \(s \in S\). From the monotonic improvement Theorem~3, this implies that
\[
\vec{v}^{\texttt{SNS}, \mu^n}(s) = \vec{v}^{\texttt{SNS}, \mu^{n+1}}(s) = \boldsymbol{Q}^{\texttt{SNS}, \mu^n}
\big(s, \mu^{n+1}(s)\big), \quad \forall s \in S.
\]
But from equations~(13) and (14), we have
\[
\boldsymbol{Q}^{\texttt{SNS}, \mu^n}\big(s, \mu^{n+1}(s)\big) = \max_{a \in A} \boldsymbol{Q}^{\texttt{SNS}, \mu^n}(s, a) = \vec{v}^{\texttt{SNS}, \mu^n}(s).
\]
This means that for all \(s \in S\),
\begin{align*}
& \vec{v}^{\texttt{SNS}, \mu^n}(s) = \max_{a \in A} \boldsymbol{Q}^{\texttt{SNS}, \mu^n}(s, a) = \max_{a \in A} \boldsymbol{r}_{\mathcal{E}}(s, a) + \gamma \max_{a \in A} \mathbb{E} \left[ \vec{v}^{\texttt{SNS}, \mu^n}(s') \,\middle|\, S_0 = s,\, A_0 = a \right].
\end{align*}
Therefore, \(\vec{v}^{\texttt{SNS}, \mu^n}\) satisfies the Bellman optimality equation. Since the optimal value function \(\vec{v}^{\texttt{SNS}}\) is the unique fixed-point of the Bellman optimality equation, we conclude that
\[
\vec{v}^{\texttt{SNS}, \mu^k}(s) = \max_{\mu} \vec{v}^{\texttt{SNS},\mu}(s),
\]
for all $s\in \mathcal{S}$. Consequently, \(\vec{v}^{\texttt{SNS}, \mu^k}\) is an optimal policy.

\section{Proof of Lemma 2}

Suppose \(\vec{Q}^{\texttt{SNS}}(s, a)\) is any function that satisfies Eq.~(17). Then the vector formed by \(\max_{a' \in \mathcal{A}} \vec{Q}^{\texttt{SNS}}(s', a')\) also satisfies Bellman's equation. By the uniqueness of Bellman solutions, it follows that
\[
\max_{a' \in \mathcal{A}} \vec{Q}^{\texttt{SNS}}(s', a') 
= \max_{a' \in \mathcal{A}} \vec{Q}^{\texttt{SNS},\star}(s', a')
\quad 
\text{for all } s' \in \mathcal{S}.
\]
Since \(\vec{Q}^{\texttt{SNS}}(s,a)\) also satisfies Eq.~(17), we conclude that \(\vec{Q}^{\texttt{SNS}}(s,a) = \vec{Q}^{\texttt{SNS},\star}(s,a)\). Hence, the solution is unique.

\section{Proof of Theorem 5}

First of all, to leverage Proposition 4.4 in \cite{bertsekas1996neuro}, we bring it here again:

\begin{proposition}\label{Thm:Bertsekas-Prop4.4}
Consider a sequence $\{u_k\}_{t=0}^{\infty}$ in $\mathbb{R}^n$ generated by a stochastic approximation algorithm of the form
\[
u_{k+1}(i) = (1 - \alpha_k(i))\,u_k(i) \;+\; \alpha_k(i)\bigl((\mathcal{T} u_k)(i) + w_k(i)\bigr), \quad i=1,\ldots,n,\; k=0,1,2,\ldots
\]
where $\{w_k(i)\}$ is a stochastic noise process and $\alpha_k(i)$ are step-sizes. Assume that $\mathcal{T}:\mathbb{R}^n \to \mathbb{R}^n$ is an operator with a fixed point $u^\star$, i.e., $\mathcal{T} u^\star = u^\star$.

\medskip
We impose the following conditions:

\begin{enumerate}
\item[\textbf{(1)}] \textbf{Step-Size Conditions:} 
   For each $i$, if $u(i)$ is not updated at time $k$, then $\alpha_k(i)=0$. The step-sizes $\{\alpha_k(i)\}$ are nonnegative and satisfy
   \[
   \sum_{k=0}^{\infty} \alpha_k(i) = \infty \quad\text{and}\quad \sum_{k=0}^{\infty} \alpha_k(i)^2 < \infty, \quad \forall i.
   \]

\item[\textbf{(2)}] \textbf{Noise Conditions:}
   Let $\mathcal{H}_k$ be the history of the algorithm up to time $k$, which is defined as follows:

   \begin{align*}
       \mathcal{H}_k = \{u_0, u_1, \cdots, u_k, w_0, w_1, \cdots, w_k, \alpha_0, \alpha_1, \cdots, \alpha_k \}
   \end{align*}
   
   Assume for all $i,k$:
   \[
   E[w_k(i) \mid \mathcal{H}_k] = 0,
   \]
   and there exist $A,B \ge 0$ such that
   \[
   E[\left(w_k(i)\right)^2 \mid \mathcal{H}_k] \le A + B\|u_k\|^2.
   \]

\item[\textbf{(3)}] \textbf{Weighted Maximum Norm Pseudo-Contraction of the Operator:}
   There exists a strictly positive vector $\xi \in \mathbb{R}^n$ (i.e., $\xi(i)>0$ for all $i$) and a constant $\beta \in [0,1)$ such that
   \[
   \|\mathcal{T} u - u^\star\|_{\xi} \le \beta \|u - u^\star\|_{\xi} \quad \forall u \in \mathbb{R}^n,
   \]
   where the weighted maximum norm is defined by
   \[
   \|u\|_{\xi} = \max_{1 \le i \le n} \frac{|u(i)|}{\xi(i)}.
   \]
\end{enumerate}

\medskip
Under the above conditions the stochastic approximation sequence $\{u_k\}$ converges almost surely to the unique fixed point $u^\star$ of $\mathcal{T}$. That is,
\[
\lim_{k \to \infty} u_k = u^\star 
\]
\end{proposition}

    We define the operator $\mathcal{T}$ as follows:
    \begin{align}\label{eq:Operator}
        (\mathcal{T}\vec{Q}^{\texttt{SNS}})(s, a) = \vec{r}_{\mathcal{E}}(s,a) + \gamma \sum_{s' \in \mathcal{S}} p(s'|s,a) \max_{a' \in \mathcal{A}} \vec{Q}^{\texttt{SNS}}(s', a'), ~~~ \forall s \in \mathcal{S}, a \in \mathcal{A}
    \end{align}

    Then, the Q-learning is defined in Eq. (15) can be shown as:
    \begin{align} \label{eq:Q-learing-algorithm}
    \vec{Q}^{\texttt{SNS}}_{k+1}(s, a) = (1-\alpha_k)\vec{Q}^{\texttt{SNS}}_{k}(s, a) + \alpha_k \left((\mathcal{T}\vec{Q}^{\texttt{SNS}}_{k})(s, a) + \mathcal{N}_k(s,a)\right)
    \end{align}
    where,
    \begin{align*}
    \mathcal{N}_k(s,a) = \vec{r}_{\mathcal{E}}(s,a) + \gamma \max_{a' \in \mathcal{A}} \vec{Q}^{\texttt{SNS}}_{k}(s', a') - (\mathcal{T}\vec{Q}^{\texttt{SNS}}_{k})(s, a)
    \end{align*}

    To demonstrate the convergence of the algorithm in Eq.~\eqref{eq:Q-learing-algorithm}, it is necessary to verify that it satisfies the conditions outlined in Proposition \eqref{Thm:Bertsekas-Prop4.4}. The step-size conditions are met by choosing appropriate values for $\alpha_k$ and adopting a suitable behavioral policy that ensures each state-action pair is visited infinitely often. For the third condition, it must be shown that the noise term $\mathcal{N}_k(\cdot)$ has zero mean and bounded variance. We can show that the noise term has zero mean as follows:

    \begin{align*}
    \mathbb{E} \left[\mathcal{N}_k(s,a) \mid \mathcal{H}_k \right] = \vec{r}_{\mathcal{E}}(s,a) + \gamma \sum_{s' \in \mathcal{S}} p(s' \mid s, a) \max_{a' \in \mathcal{A}} \vec{Q}^{\texttt{SNS}}_{k}(s', a') - (\mathcal{T}\vec{Q}^{\texttt{SNS}}_{k})(s, a) = 0
    \end{align*}

    For variance of the noise term, we have:

\begin{align*}
    \mathbb{E} \left[\left(\mathcal{N}_k(s,a)\right)^{2} \mid \mathcal{H}_k \right] &= \mathbb{E} \left[ \left(\vec{r}_{\mathcal{E}}(s,a) + \gamma \max_{a' \in \mathcal{A}} \vec{Q}^{\texttt{SNS}}_{k}(s', a') - (\mathcal{T}\vec{Q}^{\texttt{SNS}}_{k})(s, a) \right)^{2} \mid \mathcal{H}_k \right] \\
    &= \mathbb{E} \Bigg[ \left(\vec{r}_{\mathcal{E}}(s,a)\right)^{2} 
    + \gamma^{2} \left(\max_{a' \in \mathcal{A}} \vec{Q}^{\texttt{SNS}}_{k}(s', a')\right)^{2} 
    + \left((\mathcal{T}\vec{Q}^{\texttt{SNS}}_{k})(s, a)\right)^{2} \\
    & + 2\vec{r}_{\mathcal{E}}(s,a)\gamma \max_{a' \in \mathcal{A}} \vec{Q}^{\texttt{SNS}}_{k}(s', a') 
    - 2\vec{r}_{\mathcal{E}}(s,a)(\mathcal{T}\vec{Q}^{\texttt{SNS}}_{k})(s, a) \\
    & - 2\gamma \max_{a' \in \mathcal{A}} \vec{Q}^{\texttt{SNS}}_{k}(s', a')(\mathcal{T}\vec{Q}^{\texttt{SNS}}_{k})(s, a) 
    \mid \mathcal{H}_k \Bigg] 
\end{align*}

Thus, utilizing Eq. \eqref{eq:Operator}, we obtain:

\begin{align*}
    \mathbb{E} \left[\left(\mathcal{N}_k(s,a)\right)^{2} \mid \mathcal{H}_k \right] &= \left(\vec{r}_{\mathcal{E}}(s,a)\right)^{2} + \gamma^{2} \mathbb{E} \Bigg[ \left(\max_{a' \in \mathcal{A}} \vec{Q}^{\texttt{SNS}}_{k}(s', a')\right)^{2} \mid \mathcal{H}_k \Bigg]  + \left(\vec{r}_{\mathcal{E}}(s,a)\right)^{2} \\
    &+ \gamma^{2} \mathbb{E} \Bigg[ \max_{a' \in \mathcal{A}} \vec{Q}^{\texttt{SNS}}_{k}(s', a') \mid \mathcal{H}_k \Bigg]^{2} \\
    &+ 2 \gamma \vec{r}_{\mathcal{E}}(s,a) \mathbb{E} \Bigg[ \max_{a' \in \mathcal{A}} \vec{Q}^{\texttt{SNS}}_{k}(s', a') \mid \mathcal{H}_k \Bigg] 
    \\
    &+ 2\gamma \vec{r}_{\mathcal{E}}(s,a) \mathbb{E} \Bigg[\max_{a' \in \mathcal{A}} \vec{Q}^{\texttt{SNS}}_{k}(s', a') \mid \mathcal{H}_k \Bigg] - 2\left(\vec{r}_{\mathcal{E}}(s,a)\right)^{2} \\
    &- 2\gamma \vec{r}_{\mathcal{E}}(s,a) \mathbb{E} \Bigg[\max_{a' \in \mathcal{A}} \vec{Q}^{\texttt{SNS}}_{k}(s', a') \mid \mathcal{H}_k \Bigg] \\
    &-2\gamma \vec{r}_{\mathcal{E}}(s,a)\mathbb{E} \Bigg[\max_{a' \in \mathcal{A}} \vec{Q}^{\texttt{SNS}}_{k}(s', a') \mid \mathcal{H}_k \Bigg] \\
    &- 2\gamma^{2}\mathbb{E} \Bigg[ \max_{a' \in \mathcal{A}} \vec{Q}^{\texttt{SNS}}_{k}(s', a') \mid \mathcal{H}_k \Bigg]^{2} \\
    &= \gamma^{2} \mathbb{E} \Bigg[ \left(\max_{a' \in \mathcal{A}} \vec{Q}^{\texttt{SNS}}_{k}(s', a')\right)^{2} \mid \mathcal{H}_k \Bigg] \\
    &- \gamma^{2}\mathbb{E} \Bigg[ \max_{a' \in \mathcal{A}} \vec{Q}^{\texttt{SNS}}_{k}(s', a') \mid \mathcal{H}_k \Bigg]^{2} \\
    &\leq \gamma^{2} \mathbb{E} \Bigg[ \left(\max_{a' \in \mathcal{A}} \vec{Q}^{\texttt{SNS}}_{k}(s', a')\right)^{2} \mid \mathcal{H}_k \Bigg] \\
    &- \gamma^{2} \left(\underset{s' \in \mathcal{S}}{\min} \max_{a' \in \mathcal{A}} \vec{Q}^{\texttt{SNS}}_{k}(s', a') \right)^{2}
\end{align*}

Therefore, the third condition is satisfied. It remains to demonstrate that the operator is a weighted maximum norm pseudo-contraction. Before proceeding with the proof, we first highlight an interesting property of the transition probability, which will play a crucial role in the proof.
\begin{lemma} \label{lemma:transition_prob_contraction}  
    There exists a vector \(\nu\) with  positive components and a scalar \(\lambda < 1\) such that  
    \[  
    \gamma \sum_{s' \in \mathcal{S}} p(s' \mid s, a) \nu(s') \leq \lambda \nu(s),  
    \]  
    for all \(s \in \mathcal{S}\) and \(a \in \mathcal{A}\), where $\gamma \in [0,1)$.  
\end{lemma}

\begin{proof}
    We begin with Bellman's equation, under the assumption that \(\vec{r}_{\mathcal{E}}(s,a) \geq 0\) for all \(s \in \mathcal{S}\) and \(a \in \mathcal{A}\):
    \begin{align*}
    \vec{v}^{\texttt{SNS}, \star}(s) &= \max_{a \in \mathcal{A}} \vec{Q}^{\texttt{SNS},\star}(s, a) = \max_{a \in \mathcal{A}} ~ \vec{r}_{\mathcal{E}}(s,a) + \gamma \max_{a \in \mathcal{A}} ~ \sum_{s' \in \mathcal{S}} p(s'|s,a) \vec{v}^{\texttt{SNS},\star}(s') \\
    & \geq \max_{a \in \mathcal{A}} ~ \vec{r}_{\mathcal{E}}(s,a) + \gamma \sum_{s' \in \mathcal{S}} p(s'|s,a) \vec{v}^{\texttt{SNS},\star}(s')
    \end{align*}

    We define \(\nu(s)\) as \(\vec{v}^{\texttt{SNS},\star}(s)\). Thus, we have:
    \begin{align*}
    \lambda \nu(s) \geq \nu(s)-\max_{a \in \mathcal{A}}~ \vec{r}_{\mathcal{E}}(s,a) &\geq \gamma \sum_{s' \in \mathcal{S}} p(s'|s,a) \nu(s')
    \end{align*}

    where \(\lambda\) is given by:
    \begin{align*}
    \lambda = \max_{s \in \mathcal{S}} ~ \frac{\nu(s)-\max_{a \in \mathcal{A}}~ \vec{r}_{\mathcal{E}}(s,a)}{\nu(s)}  < 1
    \end{align*}
    
\end{proof}

We now utilize Lemma \ref{lemma:transition_prob_contraction} to demonstrate that the operator is a weighted maximum norm pseudo-contraction. Specifically, for any two functions \(\vec{Q}^{\texttt{SNS}}(\cdot)\) and \(\hat{\vec{Q}}^{\texttt{SNS}}(\cdot)\), and a vector \(\nu \in \mathbb{R}^{|\mathcal{S}|}\) with strictly positive elements, we can express:

\begin{align*}
    \left|(\mathcal{T}\vec{Q}^{\texttt{SNS}})(s,a)-(\mathcal{T}\hat{\vec{Q}}^{\texttt{SNS}})(s,a)\right| &= \left|\gamma \sum_{s' \in \mathcal{S}} p(s'|s,a) \max_{a' \in \mathcal{A}} \vec{Q}^{\texttt{SNS}}(s', a') - \gamma \sum_{s' \in \mathcal{S}} p(s'|s,a) \max_{a' \in \mathcal{A}} \hat{\vec{Q}}^{\texttt{SNS}}(s', a')\right| \\
    & \leq \gamma \sum_{s' \in \mathcal{S}} p(s'|s,a) \left|\max_{a' \in \mathcal{A}} \vec{Q}^{\texttt{SNS}}(s', a') - \max_{a' \in \mathcal{A}}\hat{\vec{Q}}^{\texttt{SNS}}(s', a')\right|\\
    & \leq \gamma \sum_{s' \in \mathcal{S}} p(s'|s,a) \max_{a' \in \mathcal{A}}\left| \vec{Q}^{\texttt{SNS}}(s', a') - \hat{\vec{Q}}^{\texttt{SNS}}(s', a')\right|\\
    & \leq  \| \vec{Q}^{\texttt{SNS}} - \hat{\vec{Q}}^{\texttt{SNS}}\|_{\nu} \gamma \sum_{s' \in \mathcal{S}} p(s'|s,a)  \nu(s')       ~~~~~~~ \text{(Using Lemma \ref{lemma:transition_prob_contraction})} \\
    & \leq  \lambda \| \vec{Q}^{\texttt{SNS}} - \hat{\vec{Q}}^{\texttt{SNS}}\|_{\nu} \nu(s) 
\end{align*}

We divide both sides by \(\nu(s)\) and then take the maximum over all \(s \in \mathcal{S}\) and \(a \in \mathcal{A}\), yielding:

\begin{align*}
    \|\mathcal{T}\vec{Q}^{\texttt{SNS}}-\mathcal{T}\hat{\vec{Q}}^{\texttt{SNS}}\|_{\nu} &\leq  \lambda \| \vec{Q}^{\texttt{SNS}} - \hat{\vec{Q}}^{\texttt{SNS}}\|_{\nu} 
\end{align*}

Hence, the operator $\mathcal{T}$ qualifies as a weighted maximum norm pseudo-contraction. 

To complete the convergence of the Q-learning, it needs to show that $\vec{Q}^{\texttt{SNS}}_{k}(s, a)$ is bounded. To do so, we denote $R_{max} = \max_{s \in \mathcal{S}, a \in \mathcal{A}} ~ r_{\mathcal{E}}(s,a)$. Then, it is easy to show that,

$$\vec{Q}^{\texttt{SNS}}(s, a) \leq \frac{R_{max}}{1-\gamma}, ~~~~~ \text{for all $s \in \mathcal{S}, a \in \mathcal{A}$} $$

Therefore, we have the following lemma:

\begin{lemma}
    If $\vec{Q}^{\texttt{SNS}}_0(s, a)$ is initialized such that $\vec{Q}^{\texttt{SNS}}_0(s, a) \leq \frac{R_{max}}{1-\gamma}$ for all $s \in \mathcal{S}$ and $a \in \mathcal{A}$, then $\vec{Q}^{\texttt{SNS}}_{k'}(s, a)$ remains bounded by $\frac{R_{max}}{1-\gamma}$ for all $s \in \mathcal{S}$, $a \in \mathcal{A}$, and $k' \geq 0$.
\end{lemma}

\begin{proof}
The proof proceeds by induction. For all \( s \in \mathcal{S} \) and \( a \in \mathcal{A} \), it holds that \( \vec{Q}^{\texttt{SNS}}_0(s, a) \leq \frac{R_{\text{max}}}{1 - \gamma} \). Consequently, using Eq.~\eqref{eq:Q-learing-algorithm}, we have:

\begin{align*}
    \vec{Q}^{\texttt{SNS}}_{1}(s, a) &= (1-\alpha_0)\vec{Q}^{\texttt{SNS}}_{0}(s, a) + \alpha_0 \left(\vec{r}_{\mathcal{E}}(s,a) + \gamma \max_{a' \in \mathcal{A}} \vec{Q}^{\texttt{SNS}, \mu}_{0}(s', a')\right) \\
    & = (1-\alpha_0)\frac{R_{max}}{1-\gamma} + \alpha_0 \left(R_{max} + \gamma \frac{R_{max}}{1-\gamma}\right) = \left( (1-\alpha_0) + \alpha_0 \left((1-\gamma) + \gamma \right) \right) \frac{R_{max}}{1-\gamma} \\
    &= \frac{R_{max}}{1-\gamma}
\end{align*}

Thus, assuming that $\vec{Q}^{\texttt{SNS}}_k(s, a) \leq \frac{R_{max}}{1-\gamma}$ holds true for all $s \in \mathcal{S}$ and $a \in \mathcal{A}$, we can express:

\begin{align*}
    \vec{Q}^{\texttt{SNS}}_{k+1}(s, a) &= (1-\alpha_k)\vec{Q}^{\texttt{SNS}}_{k}(s, a) + \alpha_k \left(\vec{r}_{\mathcal{E}}(s,a) + \gamma \max_{a' \in \mathcal{A}} \vec{Q}^{\texttt{SNS}}_{k}(s', a')\right) \\
    & = (1-\alpha_k)\frac{R_{max}}{1-\gamma} + \alpha_k \left(R_{max} + \gamma \frac{R_{max}}{1-\gamma}\right) = \left( (1-\alpha_k) + \alpha_k \left((1-\gamma) + \gamma \right) \right) \frac{R_{max}}{1-\gamma} \\
    &= \frac{R_{max}}{1-\gamma}
\end{align*}

In conclusion, $\vec{Q}^{\texttt{SNS}}_{k'}(s, a)$ is guaranteed to remain bounded by $\frac{R_{max}}{1-\gamma}$ for all $s \in \mathcal{S}$, $a \in \mathcal{A}$, and $k' \geq 0$.

\end{proof}

Consequently, since all the required conditions are met and $\vec{Q}^{\texttt{SNS}}(\cdot)$ is bounded, it follows that $\vec{Q}^{\texttt{SNS}}(\cdot)$ converges almost surely to the unique fixed point $\vec{Q}^{\texttt{SNS},\star}(\cdot)$ of $\mathcal{T}$.

\section{Markov Chains and Markov Reward Processes} \label{Section:MarkovChains}

In this section, we review some relevant background on Markov Chains and Markov Reward Processes (MRPs) that are needed for our proofs and results in the paper.
%
%
%
\subsection{Markov Chains} \label{SubSection:P_of_MCs}

This section begins with an introduction to the essential characteristics of Markov chains, as described in Chapter 1 of~\cite{levin2017markov}. A Markov chain is defined as a pair \((\mathcal{S}, p(\cdot))\), where \(\mathcal{S}\) represents a finite set of states and \(p(\cdot)\) is the transition function. Specifically, $$p: \mathcal{S} \times \mathcal{S} \rightarrow [0,1]$$ represents the probability function for state transitions, with \(p(s'|s)\) indicating the probability of moving from state \(s\) to state \(s'\). For each state \(s\), it holds that \(p(s'|s) \geq 0\) for every \(s' \in \mathcal{S}\) and 
    $$ \sum_{s' \in \mathcal{S}} p(s'|s) = 1.$$
We also utilize a matrix representation for the Markov chain transitions, denoted \(P \in \mathbb{R}^{n \times n}\), where 
    $$P(s, s') = p(s'|s).$$
The progression of states in a Markov chain is depicted by a sequence of random variables:
 \begin{align} \label{eq:trajectory} S_0, S_1, \ldots, S_k, \ldots, \end{align}
with the transition probability from state \(S_k = s\) to \(S_{k+1} = s'\) given by \(\Pr{S_{k+1} = s'|S_k = s} = p(s'|s)\). For any \(t \in \mathbb{N}\),
$$p^t(s'|s) := \Pr{S_{k+t} = s'|S_k = s},$$
denotes the transition probability to state \(s'\) after \(t\) steps starting from state \(s\), and can be calculated as
$$p^k(s'|s) = P^k(s', s).$$

A Markov chain \((\mathcal{S}, p(\cdot))\) is termed \emph{irreducible} if for any two states \(s, s'\) there exists a \(k \in \mathbb{N}\) such that \(p^k(s'|s) > 0\). For any state \(s\), define
$$ \mathcal{T}(s) = \{t \geq 1 | p^t(s, s) > 0\}.$$
The \emph{period} of a state \(s\) is the greatest common divisor of the set \(\mathcal{T}(s)\). If the Markov chain is irreducible, all states share the same period, referred to as the chain's period. A chain is \emph{aperiodic} if every state has a period of 1. The following propositions are useful \cite{levin2017markov}:
\begin{proposition} \label{Prop:invariantDist}
If \((\mathcal{S}, p(\cdot))\) is both irreducible and aperiodic, then there exists a unique distribution, \(\vecpi \in \mathbb{R}^{|\mathcal{S}|}\), such that \(\vecpi(s) > 0\) for every \(s \in \mathcal{S}\), and 
$$\sum_{s \in \mathcal{S}} \vecpi(s) = 1, \quad \text{and} \quad \vecpi = \vec{P}^{\rm{T}} \vecpi.$$
Furthermore, for each \(s, s'\) in \(\mathcal{S}\),
$$\vecpi(s') = \lim_{k \rightarrow \infty} p^k(s'|s).$$
This distribution is referred to as the \emph{invariant distribution} of the Markov chain.
\end{proposition}
\begin{proposition} \label{Prop:MCequivalence}
A Markov chain \((\mathcal{S}, p(\cdot))\) is irreducible and aperiodic if and only if there is a \(K \in \mathbb{N}\) such that for all \(s, s' \in \mathcal{S}\) and for all \(k \geq K\),
$$p^k(s'|s) > 0.$$
\end{proposition}

\subsection{Markov Reward Process (MRP)}

A Markov Reward Process (MRP) is defined as a tuple \(M = (\mathcal{S}, p(\cdot), \vec{r}, \gamma)\). The set \(\mathcal{S}\) represents a finite state space, \(p(\cdot)\) is the transition function of the Markov chain, \(\vec{r} \in \mathbb{R}^{|\mathcal{S}|}\) denotes a reward vector where \(\vec{r}(s)\) signifies the immediate reward for being in state \(s\), and \(\gamma\) is a discount factor that quantifies the relative importance of immediate versus future rewards. The dynamics of an MRP are captured by a sequence of state-reward pairs, represented by the sequence of random variables \(S_0, R_0, S_1, R_1, \ldots, S_k, R_k, \ldots\), where \(k \in \mathbb{N}\) is a time index, and \(R_k = \vec{r}(S_k)\) is the reward received at time \(k\).

Value estimation is a primary task in studying MRPs, focusing on determining the value function from each state. This value function is denoted by the vector \(\vec{v} \in \mathbb{R}^{|\mathcal{S}|}\), and is defined as
$$\vec{v}(s) = \mathbb{E} \left[ \sum_{k=0}^{\infty} \gamma^k R_k \mid S_0 = s \right].$$

According to the paper in \cite{tsitsiklis1997analysis}, the vector \(\vec{v}\) can be calculated using the formula
$$\vec{v} = (\vec{I} - \gamma \vec{P})^{-1} \vec{r}$$
which relies on both the reward vector and the transition matrix \(\vec{P}\). However, in many practical situations, the exact transition probabilities and rewards are unknown, and analysts must rely on data from sampled trajectories as depicted in~\eqref{eq:trajectory}.

\subsection{Temporal Difference (TD) Learning}

Temporal Difference (TD) Learning is known as an effective stochastic approach for estimating the value vector $\vec{v}$ via a sample trajectory. This methodology utilizes the Temporal Difference evaluation algorithm, which progressively refines an estimation $\vec{v}_k \in \mathbb{R}^n$ of $\vec{v}$. Each iteration involves updating the estimate based on each sample $(S_k, R_k, S_{k+1})$ from the MRP trajectory, starting from any initial condition $\vec{v}_0 \in \mathbb{R}^{|\mathcal{S}|}$. For each step $k$, the next estimate $\vec{v}_{k+1}$ is calculated as follows:
\begin{align*}
    \vec{v}_{k+1}(s) = 
     \begin{cases}
       \vec{v}_k(s) + \alpha_k (R_k + \gamma \vec{v}_k(S_{k+1}) - \vec{v}_k(s)) & \text{if } s = S_k \\
      \vec{v}_k(s) & \text{if } s \neq S_k
     \end{cases}
\end{align*}
where $\alpha_k > 0$ is a positive step-size. The convergence of this iterative process to the true value function is contingent upon the proper selection of step sizes as follows: 
\begin{assumption}\label{assumption:StepSizes}
    The step sizes $\alpha_k$ are deterministic, non-negative, and meet the following criteria:
\begin{equation}\label{Equ:E1supp}
    \sum_{k=0}^{\infty} \alpha_k = \infty \quad \text{and} \quad \sum_{k=0}^{\infty} \alpha_k^2 < \infty.
\end{equation}
\end{assumption}

Under these conditions, and given that the MRP characterized by $(\mathcal{S}, p(\cdot))$ is irreducible and aperiodic, it is established that the TD algorithm converges to this theoretical fixed point almost surely, as represented by \cite{levin2017markov}:
\begin{align}\label{eq:TD_fixedpoint}
    \lim_{k \rightarrow \infty} \vec{v}_k = \vec{v} = (\vec{I} - \gamma \vec{P})^{-1} \vec{r}.
\end{align}

\section{Details on the Experiments}

We demonstrated our theoretical results in the context of wireless communication systems in Section 9, which frequently experience dynamic channel conditions due to factors such as fading, interference, and user mobility.

Wireless communication systems are inherently dynamic and complex because of the unpredictable nature of the wireless medium, which causes the quality of the wireless channel to fluctuate over time and across different locations. This variability is influenced by several factors. One is fading, fluctuations in signal strength caused by the constructive and destructive interference of multiple signal paths. Another is interference, unwanted signals from other transmitters that disrupt communication. User mobility also plays a role, as the movement of users alters signal propagation conditions.

To enhance performance under such fluctuating conditions, Adaptive Modulation (AM) techniques are employed. Adaptive Modulation involves dynamically adjusting transmission parameters, such as modulation schemes, to match current channel conditions~\cite{huang2020adaptive,qiu1999performance}. This approach aims to maximize data throughput while maintaining reliable communication.

To showcase the effectiveness of our proposed framework, we modeled an adaptive communication system using the SNS-MDP framework. The SNS-MDP effectively captures the stochastic and time-varying nature of wireless environments.

In our model, the transceiver functions as an agent that makes decisions based on observations of the system state. Specifically, the agent selects a frequency band for data transmission after observing the current modulation scheme.
\[
\mathcal{A} = \{ \texttt{FB}_1, \texttt{FB}_2, \ldots, \texttt{FB}_A \},
\]
where $\texttt{FB}_i$ represents the $i$-th frequency band, and $A$ is the total number of available frequency bands.

The states in the system correspond to different Modulation Schemes (MS), each offering a unique trade-off between data rate and noise tolerance:
\[
\mathcal{S} = \{ \texttt{MS}_1, \texttt{MS}_2, \ldots, \texttt{MS}_S \},
\]
where $\texttt{MS}_j$ represents the $j$-th modulation scheme, and $S$ is the number of available modulation schemes.

The environmental states represent the channel conditions, which are crucial yet typically unobservable factors that influence communication dynamics.
\[
\mathcal{E} = \{ \text{Excellent (E)}, \text{Good (G)}, \text{Fair (F)}, \text{Poor (P)} \}.
\]

\subsection{Markovian Dynamics of Channel Conditions}

Channel conditions are often modeled using Markovian dynamics, with transitions governed by a probability matrix $q(e'|e)$ \cite{sanchez2007n}. This approach captures the temporal dependencies of channel conditions due to factors like fading and mobility. Channel condition transition probability can be estimated, but in this paper, we just use some predefined values to show the convergence of the RL algorithms upon the SNS-MDP framework. Table \ref{tab:env_setting} shows the content of channel condition transition probability \footnote{All the values for the probabilities in the Tables 
 are scaled from 0 to 1.}. 

\begin{table}[h]
\centering
\caption{Environment Setting Transition Probabilities}
\label{tab:env_setting}
\begin{tabular}{lcccc}
\hline
& \multicolumn{4}{c}{Next State} \\
Current State & Excellent & Good & Fair & Poor \\
\hline
Excellent & 0.44 & 0.11 & 0.12 & 0.33 \\
Good      & 0.20 & 0.10 & 0.30 & 0.40 \\
Fair      & 0.66 & 0.11 & 0.09 & 0.14 \\
Poor      & 0.18 & 0.22 & 0.40 & 0.20 \\
\hline
\end{tabular}

\end{table}

In practice, {Probability of Successful Transmission}, which is denoted by $P_{\text{success}}(s, e, a)$ can be estimated through empirical measurements or analytical models~\cite{pan2021success,weber2010overview}. For our simulation, we use predefined values to focus on demonstrating the convergence properties of our algorithms~\cite{halloush2022formula}. In Table \ref{table:Psuccess_grp1} and \ref{table:Psuccess_grp2}, there are the detailed values for each \( P_{\text{success}}(s, e, a) \). Once a frequency band is selected, the corresponding table is chosen, where each table contains the probability of successful transmission for each pair of modulation schemes and channel conditions.

\begin{table}[h]
\centering
\caption{Probability of Successful Transmission in Frequency Bands 1 to 5}
\label{table:Psuccess_grp1}
\begin{tabular}{cc}
\begin{minipage}{0.48\linewidth}
\centering
\textbf{Frequency Band 1}
\begin{tabular}{lcccc}
\hline
\textbf{MS} & \textbf{Excellent} & \textbf{Good} & \textbf{Fair} & \textbf{Poor} \\
\hline
BPSK       & 0.83 & 0.84 & 0.89 & 0.86 \\
QPSK       & 0.99 & 0.78 & 0.80 & 0.79 \\
8-PSK      & 0.91 & 0.81 & 0.87 & 0.81 \\
16-QAM     & 0.79 & 0.78 & 0.91 & 0.78 \\
32-QAM     & 0.88 & 0.81 & 0.88 & 0.75 \\
64-QAM     & 0.92 & 0.85 & 0.84 & 0.72 \\
128-QAM    & 0.87 & 0.80 & 0.83 & 0.74 \\
256-QAM    & 0.91 & 0.82 & 0.86 & 0.70 \\
512-QAM    & 0.93 & 0.86 & 0.90 & 0.68 \\
1024-QAM   & 0.85 & 0.79 & 0.81 & 0.71 \\
2048-QAM   & 0.89 & 0.83 & 0.84 & 0.69 \\
\hline
\end{tabular}
\end{minipage}
&
\begin{minipage}{0.48\linewidth}
\centering
\textbf{Frequency Band 2}
\begin{tabular}{lcccc}
\hline
\textbf{MS} & \textbf{Excellent} & \textbf{Good} & \textbf{Fair} & \textbf{Poor} \\
\hline
BPSK       & 0.72 & 0.84 & 0.89 & 0.83 \\
QPSK       & 0.94 & 0.87 & 0.67 & 0.66 \\
8-PSK      & 0.78 & 0.79 & 0.72 & 0.72 \\
16-QAM     & 0.74 & 0.71 & 0.93 & 0.73 \\
32-QAM     & 0.79 & 0.75 & 0.87 & 0.71 \\
64-QAM     & 0.81 & 0.77 & 0.85 & 0.70 \\
128-QAM    & 0.82 & 0.78 & 0.86 & 0.69 \\
256-QAM    & 0.85 & 0.80 & 0.88 & 0.68 \\
512-QAM    & 0.83 & 0.81 & 0.84 & 0.67 \\
1024-QAM   & 0.88 & 0.83 & 0.82 & 0.65 \\
2048-QAM   & 0.86 & 0.85 & 0.80 & 0.64 \\
\hline
\end{tabular}
\end{minipage}
\\[2ex]
\begin{minipage}{0.48\linewidth}
\centering
\textbf{Frequency Band 3}
\begin{tabular}{lcccc}
\hline
\textbf{MS} & \textbf{Excellent} & \textbf{Good} & \textbf{Fair} & \textbf{Poor} \\
\hline
BPSK       & 0.56 & 0.61 & 0.83 & 0.68 \\
QPSK       & 0.82 & 0.81 & 0.88 & 0.65 \\
8-PSK      & 0.83 & 0.81 & 0.61 & 0.61 \\
16-QAM     & 0.63 & 0.86 & 0.59 & 0.89 \\
32-QAM     & 0.68 & 0.82 & 0.64 & 0.71 \\
64-QAM     & 0.72 & 0.83 & 0.65 & 0.73 \\
128-QAM    & 0.74 & 0.84 & 0.66 & 0.75 \\
256-QAM    & 0.76 & 0.85 & 0.67 & 0.77 \\
512-QAM    & 0.78 & 0.86 & 0.68 & 0.79 \\
1024-QAM   & 0.80 & 0.87 & 0.69 & 0.81 \\
2048-QAM   & 0.82 & 0.88 & 0.70 & 0.83 \\
\hline
\end{tabular}
\end{minipage}
&
\begin{minipage}{0.48\linewidth}
\centering
\textbf{Frequency Band 4}
\begin{tabular}{lcccc}
\hline
\textbf{MS} & \textbf{Excellent} & \textbf{Good} & \textbf{Fair} & \textbf{Poor} \\
\hline
BPSK       & 0.088 & 0.088 & 0.091 & 0.081 \\
QPSK       & 0.089 & 0.094 & 0.083 & 0.096 \\
8-PSK      & 0.094 & 0.091 & 0.096 & 0.096 \\
16-QAM     & 0.086 & 0.084 & 0.084 & 0.085 \\
32-QAM     & 0.091 & 0.087 & 0.088 & 0.086 \\
64-QAM     & 0.092 & 0.089 & 0.089 & 0.087 \\
128-QAM    & 0.093 & 0.090 & 0.090 & 0.088 \\
256-QAM    & 0.094 & 0.091 & 0.091 & 0.089 \\
512-QAM    & 0.095 & 0.092 & 0.092 & 0.090 \\
1024-QAM   & 0.096 & 0.093 & 0.093 & 0.091 \\
2048-QAM   & 0.097 & 0.094 & 0.094 & 0.092 \\
\hline
\end{tabular}
\end{minipage}
\\[2ex]
\begin{minipage}{0.48\linewidth}
\centering
\textbf{Frequency Band 5}
\begin{tabular}{lcccc}
\hline
\textbf{MS} & \textbf{Excellent} & \textbf{Good} & \textbf{Fair} & \textbf{Poor} \\
\hline
BPSK       & 0.0070 & 0.0070 & 0.0060 & 0.0010 \\
QPSK       & 0.0075 & 0.0073 & 0.0065 & 0.0020 \\
8-PSK      & 0.0080 & 0.0079 & 0.0067 & 0.0040 \\
16-QAM     & 0.0082 & 0.0081 & 0.0076 & 0.0064 \\
32-QAM     & 0.0089 & 0.0082 & 0.0078 & 0.0063 \\
64-QAM     & 0.0091 & 0.0084 & 0.0080 & 0.0062 \\
128-QAM    & 0.0090 & 0.0086 & 0.0082 & 0.0061 \\
256-QAM    & 0.0093 & 0.0088 & 0.0083 & 0.0060 \\
512-QAM    & 0.0092 & 0.0087 & 0.0084 & 0.0059 \\
1024-QAM   & 0.0095 & 0.0089 & 0.0085 & 0.0058 \\
2048-QAM   & 0.0096 & 0.0091 & 0.0086 & 0.0057 \\
\hline
\end{tabular}
\end{minipage}
&
\\
\end{tabular}
\end{table}

\begin{table}[h]
\centering
\caption{Probability of Successful Transmission in Frequency Bands 6 to 11}
\label{table:Psuccess_grp2}
\begin{tabular}{cc}
\begin{minipage}{0.48\linewidth}
\centering
\textbf{Frequency Band 6}
\begin{tabular}{lcccc}
\hline
\textbf{MS} & \textbf{Excellent} & \textbf{Good} & \textbf{Fair} & \textbf{Poor} \\
\hline
BPSK       & 0.79 & 0.81 & 0.76 & 0.67 \\
QPSK       & 0.88 & 0.82 & 0.78 & 0.66 \\
8-PSK      & 0.85 & 0.84 & 0.79 & 0.65 \\
16-QAM     & 0.90 & 0.85 & 0.80 & 0.64 \\
32-QAM     & 0.92 & 0.87 & 0.81 & 0.63 \\
64-QAM     & 0.93 & 0.88 & 0.82 & 0.62 \\
128-QAM    & 0.95 & 0.89 & 0.83 & 0.61 \\
256-QAM    & 0.94 & 0.90 & 0.84 & 0.60 \\
512-QAM    & 0.96 & 0.91 & 0.85 & 0.59 \\
1024-QAM   & 0.97 & 0.92 & 0.86 & 0.58 \\
2048-QAM   & 0.98 & 0.93 & 0.87 & 0.57 \\
\hline
\end{tabular}
\end{minipage}
&
\begin{minipage}{0.48\linewidth}
\centering
\textbf{Frequency Band 7}
\begin{tabular}{lcccc}
\hline
\textbf{MS} & \textbf{Excellent} & \textbf{Good} & \textbf{Fair} & \textbf{Poor} \\
\hline
BPSK       & 0.82 & 0.80 & 0.74 & 0.066 \\
QPSK       & 0.87 & 0.82 & 0.76 & 0.065 \\
8-PSK      & 0.89 & 0.84 & 0.77 & 0.064 \\
16-QAM     & 0.91 & 0.85 & 0.78 & 0.063 \\
32-QAM     & 0.93 & 0.87 & 0.79 & 0.062 \\
64-QAM     & 0.94 & 0.88 & 0.80 & 0.061 \\
128-QAM    & 0.95 & 0.89 & 0.81 & 0.060 \\
256-QAM    & 0.96 & 0.90 & 0.82 & 0.059 \\
512-QAM    & 0.97 & 0.91 & 0.83 & 0.058 \\
1024-QAM   & 0.98 & 0.92 & 0.84 & 0.057 \\
2048-QAM   & 0.99 & 0.93 & 0.85 & 0.0056 \\
\hline
\end{tabular}
\end{minipage}
\\[2ex]
\begin{minipage}{0.48\linewidth}
\centering
\textbf{Frequency Band 8}
\begin{tabular}{lcccc}
\hline
\textbf{MS} & \textbf{Excellent} & \textbf{Good} & \textbf{Fair} & \textbf{Poor} \\
\hline
BPSK       & 0.85 & 0.82 & 0.78 & 0.65 \\
QPSK       & 0.89 & 0.84 & 0.79 & 0.64 \\
8-PSK      & 0.92 & 0.86 & 0.80 & 0.63 \\
16-QAM     & 0.93 & 0.87 & 0.81 & 0.62 \\
32-QAM     & 0.94 & 0.88 & 0.82 & 0.61 \\
64-QAM     & 0.95 & 0.89 & 0.83 & 0.60 \\
128-QAM    & 0.96 & 0.90 & 0.84 & 0.59 \\
256-QAM    & 0.97 & 0.91 & 0.85 & 0.58 \\
512-QAM    & 0.98 & 0.92 & 0.86 & 0.57 \\
1024-QAM   & 0.99 & 0.93 & 0.87 & 0.56 \\
2048-QAM   & 1.00 & 0.94 & 0.88 & 0.55 \\
\hline
\end{tabular}
\end{minipage}
&
\begin{minipage}{0.48\linewidth}
\centering
\textbf{Frequency Band 9}
\begin{tabular}{lcccc}
\hline
\textbf{MS} & \textbf{Excellent} & \textbf{Good} & \textbf{Fair} & \textbf{Poor} \\
\hline
BPSK       & 0.88 & 0.84 & 0.80 & 0.64 \\
QPSK       & 0.92 & 0.85 & 0.81 & 0.63 \\
8-PSK      & 0.93 & 0.86 & 0.82 & 0.62 \\
16-QAM     & 0.95 & 0.87 & 0.83 & 0.61 \\
32-QAM     & 0.96 & 0.88 & 0.84 & 0.60 \\
64-QAM     & 0.97 & 0.89 & 0.85 & 0.59 \\
128-QAM    & 0.98 & 0.90 & 0.86 & 0.58 \\
256-QAM    & 0.99 & 0.91 & 0.87 & 0.57 \\
512-QAM    & 1.00 & 0.92 & 0.88 & 0.56 \\
1024-QAM   & 0.99 & 0.93 & 0.89 & 0.55 \\
2048-QAM   & 0.98 & 0.94 & 0.90 & 0.54 \\
\hline
\end{tabular}
\end{minipage}
\\[2ex]
\begin{minipage}{0.48\linewidth}
\centering
\textbf{Frequency Band 10}
\begin{tabular}{lcccc}
\hline
\textbf{MS} & \textbf{Excellent} & \textbf{Good} & \textbf{Fair} & \textbf{Poor} \\
\hline
BPSK       & 0.90 & 0.85 & 0.82 & 0.63 \\
QPSK       & 0.93 & 0.86 & 0.83 & 0.62 \\
8-PSK      & 0.94 & 0.87 & 0.84 & 0.61 \\
16-QAM     & 0.96 & 0.88 & 0.85 & 0.60 \\
32-QAM     & 0.97 & 0.89 & 0.86 & 0.59 \\
64-QAM     & 0.98 & 0.90 & 0.87 & 0.58 \\
128-QAM    & 0.99 & 0.91 & 0.88 & 0.57 \\
256-QAM    & 1.00 & 0.92 & 0.89 & 0.56 \\
512-QAM    & 0.99 & 0.93 & 0.90 & 0.55 \\
1024-QAM   & 0.98 & 0.94 & 0.91 & 0.54 \\
2048-QAM   & 0.97 & 0.95 & 0.92 & 0.53 \\
\hline
\end{tabular}
\end{minipage}
&
\begin{minipage}{0.48\linewidth}
\centering
\textbf{Frequency Band 11}
\begin{tabular}{lcccc}
\hline
\textbf{MS} & \textbf{Excellent} & \textbf{Good} & \textbf{Fair} & \textbf{Poor} \\
\hline
BPSK       & 0.91 & 0.87 & 0.84 & 0.62 \\
QPSK       & 0.94 & 0.88 & 0.85 & 0.61 \\
8-PSK      & 0.95 & 0.89 & 0.86 & 0.60 \\
16-QAM     & 0.97 & 0.90 & 0.87 & 0.59 \\
32-QAM     & 0.98 & 0.91 & 0.88 & 0.58 \\
64-QAM     & 0.99 & 0.92 & 0.89 & 0.57 \\
128-QAM    & 1.00 & 0.93 & 0.90 & 0.56 \\
256-QAM    & 0.99 & 0.94 & 0.91 & 0.55 \\
512-QAM    & 0.98 & 0.95 & 0.92 & 0.54 \\
1024-QAM   & 0.97 & 0.96 & 0.93 & 0.53 \\
2048-QAM   & 0.96 & 0.97 & 0.94 & 0.52 \\
\hline
\end{tabular}
\end{minipage}
\\
\end{tabular}
\end{table}

The {state transition probabilities} $p_e(s'|s, a)$ are influenced by $P_{\text{success}}(s, e, a)$ and are defined as:

\[
p_e(s'|s, a) = \begin{cases}
P_{\text{success}}(s, e, a), & \text{if } s' = s, \\
\displaystyle \frac{1 - P_{\text{success}}(s, e, a)}{\text{Index}(s') \times \sum_{k=1}^{|\mathcal{S}| - 1}\frac{1}{k}}, & \text{if } s' \neq s,
\end{cases}
\]

where $\text{Index}(s')$ returns the position of modulation scheme $s'$ in the ordered list starting from 1. This formulation ensures that if transmission is successful, the state remains the same; otherwise, it transitions to other states with a probability inversely proportional to their indices.

 The reward function $\vec{R}(s, e)$ measures system performance by balancing data throughput with penalties for unfavorable conditions, expressed as: 
\[
\vec{R}(s, e) = \alpha \cdot \text{Rate}(s) \cdot \text{Decay}(e) - \beta \cdot \text{Decay}(e),
\]

where $\alpha$ controls the importance of the data rate, $\beta$ penalizes the use of higher-order schemes in poor conditions, $\text{Rate}(s)$ is the data rate linked to modulation scheme $s$, and $\text{Decay}(e)$ represents degradation due to channel condition $e$. This formulation promotes modulation schemes that maximize throughput while discouraging risky decisions under poor conditions. In the simulation, $\alpha$ set to 10 and $\beta$ set to 2. Table \ref{table:datarates} and \ref{table:decayrates} represent the content of the date rate for each modulation scheme and the decay rate for each channel condition.

\begin{table}[h]
\centering
\begin{minipage}{0.45\textwidth}
    \centering
    \caption{Data Rates for Different Modulation Schemes}
    \label{table:datarates}
    \begin{tabular}{l c}
    \hline
    \textbf{MS} & \textbf{Data Rate} \\
    \hline
    BPSK       & 10  \\
    QPSK       & 20  \\
    8-PSK      & 30  \\
    16-QAM     & 40  \\
    32-QAM     & 50  \\
    64-QAM     & 60  \\
    128-QAM    & 70  \\
    256-QAM    & 80  \\
    512-QAM    & 90  \\
    1024-QAM   & 100 \\
    2048-QAM   & 110 \\
    \hline
    \end{tabular}
    
\end{minipage}%
\hfill
\begin{minipage}{0.45\textwidth}
    \centering
    \caption{Decay Rates for Different Channel Conditions}
    \label{table:decayrates}
    \begin{tabular}{l c}
    \hline
    \textbf{Channel Condition} & \textbf{Decay Rate} \\
    \hline
    Excellent & 0.99 \\
    Good      & 0.70 \\
    Fair      & 0.50 \\
    Poor      & 0.30 \\
    \hline
    \end{tabular}
    
\end{minipage}
\end{table}

The simulations are done in Python code, which is available through the link below
\footnote{\url{https://github.com/mohsen1amiri/SNS-MDP}.}. All the algorithms start with the same initial policy that recommends frequency band 1 for all the modulation schemes. The results are shown in Section 9 of the paper.

\clearpage


\end{document}